\documentclass[a4paper,twocolumn]{article}
\usepackage[width=17cm,height=22cm]{geometry}
\usepackage[english]{babel}
\usepackage[utf8]{inputenc}
\usepackage{fancyvrb}
\usepackage{authblk}
\usepackage{graphicx}
\usepackage{color}
\usepackage{xcolor}
\usepackage{hyperref}

\title{\hspace{0.5cm}\\[-2cm]\textbf{Using an expert deviation carrying the knowledge of climate data in usual clustering algorithms}}
\author[1,2]{Emmanuel Biabiany\thanks{Corresponding author. Tel: +33 590 483 110\\Email address: \href{mailto:emmanuel.biabiany@univ-antilles.fr}{emmanuel.biabiany@univ-antilles.fr}}}
\author[2]{Vincent Page}
\author[1]{Didier Bernard}
\author[2]{Hélène Paugam-Moisy}
\affil[1]{\normalsize Laboratoire de Recherche en Géosciences et Energie (LARGE), Université des Antilles}
\affil[2]{\normalsize Laboratoire de Mathématique, Informatique et Applications (LAMIA), Université des Antilles}
\date{\small \raggedright Accepted: 18 May 2020 / Joint Conferences Cap and RFIAP 2020, \textit{Vannes (France)}}

\begin{document}
\maketitle

\begin{abstract}
In order to help physicists to expand their knowledge of the climate in the Lesser Antilles, we aim to identify the spatio-temporal configurations using clustering analysis on wind speed and cumulative rainfall datasets. But we show that using the L2 norm in conventional clustering methods as K-Means (KMS) and Hierarchical Agglomerative Clustering (HAC) can induce undesirable effects. So, we propose to replace Euclidean distance (L2) by a dissimilarity measure named Expert Deviation (ED). Based on the symmetrized Kullback-Leibler divergence, the ED integrates the properties of the observed physical parameters and climate knowledge. This measure helps comparing histograms of four patches, corresponding to geographical zones, that are influenced by atmospheric structures. The combined evaluation of the internal homogeneity and the separation of the clusters obtained using ED and L2 was performed. The results, which are compared using the silhouette index, show five clusters with high indexes. For the two available datasets one can see that, unlike KMS-L2, KMS-ED discriminates the daily situations favorably, giving more physical meaning to the clusters discovered by the algorithm. The effect of patches is observed in the spatial analysis of representative elements for KMS-ED. The ED is able to produce different configurations which makes the usual atmospheric structures clearly identifiable. Atmospheric physicists can interpret the locations of the impact of each cluster on a specific zone according to atmospheric structures. KMS-L2 does not lead to such an interpretability, because the situations represented are spatially quite smooth. This climatological study illustrates the advantage of using ED as a new approach.
\end{abstract}

\medskip

\noindent\textbf{Keywords}: Clustering; Climate data; Kullback-Leibler; Silhouette index.

\section{Introduction}
\label{sec:intro}
Climate data are spatio-temporal in nature and handled by complex dynamics. In order to analyze and extract knowledge from them, machine learning methods are welcomed \cite{article59, article5,article34}. A research domain named ``Climate Informatics'' covers the subject of such an approach \cite{article4}. Among the machine learning methods, clustering algorithms are applied in the present study. Unsupervised classification methods such as K-means (KMS) or Hierarchical Agglomerative Clustering (HAC) should allow data with similar spatial patterns to be grouped together so that global trends may be identified within clusters.  Surprisingly, the literature shows no numerical evaluation of the quality of the clusters and this evaluation relies exclusively on visual inspection by experts.
 
\noindent In this article, we show that for data describing meteorological parameters, the most common clustering algorithms, such as KMS and HAC with default settings produce clusters with no physical relevance. It is also shown that the origin of this weakness mainly lies in the use of Euclidean distance (L2) as a measure of the dissimilarity between two weather patterns represented by vectors of nearly 20,000 components. We propose a new measure, called Expert Deviation (ED), based on an Image Retrieval approach, combined to a physical expertise and the use of Kullback-Leibler (KL) symmetrized divergence \cite{article68,article78,article67}. ED is used in place of L2 in the clustering methods, in order to get more physical relevance and better built clusters.

\noindent Moreover, coherence of the clusters will be assessed by silhouette index to evaluate the internal homogeneity of each cluster and the separation between them \cite{article29,Boltz2007}. The silhouette index also provides information in the selection of the number of clusters and the algorithm to be retained from the clustering analysis.

\noindent Two examples of applications are presented to describe the conception and the use of an ED for a meteorological parameter: ED$_{RAINFALL}$ for daily cumulative rainfall and ED$_{WIND}$ for daily mean wind speed.
The datasets used and the methodology developed are presented in Section \ref{sec:mat_meth}. Section \ref{sec:results} provides an explanation and evaluation of the results obtained for the study. The interpretation of these results is discussed in Section \ref{sec:discuss}. Section \ref{sec:clc_pers} concludes the article and gives some perspectives.

\section{Materials and methods}
\label{sec:mat_meth}
\subsection{Datasets}
\label{subsec:data}
\begin{figure}[hbt!]
\centering
\includegraphics[scale=0.18]{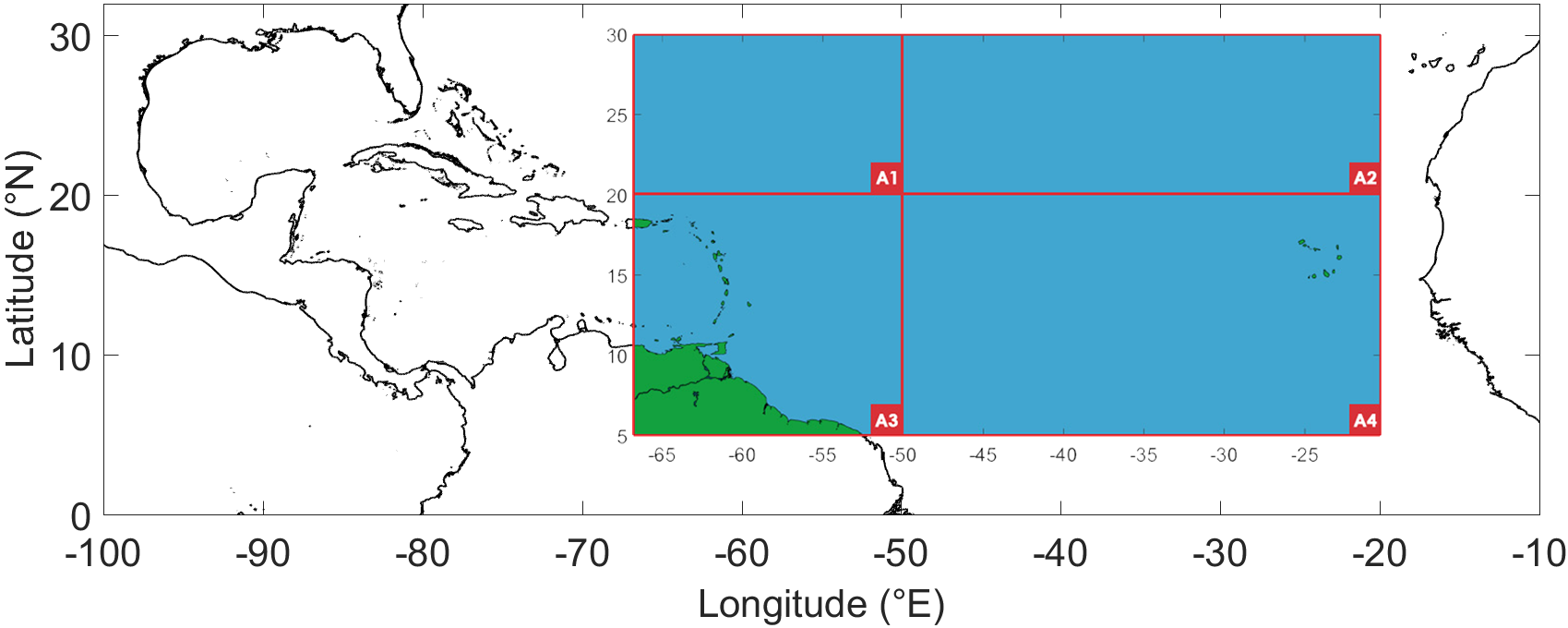}
\caption{Area of interest. Land is in zone A3 (bottom left): Lesser Antilles with a northeasterly part of South America. Zones A1, A2 and A4 are mainly sea: a part of the Central Atlantic Ocean and the Cape Verde archipelago.}
\label{fig:sch_domain}
\end{figure}

The first set of data comes from the reanalysis daily outputs of the ECMWF ERA-5 model for the wind components at an 850 hPa isobar (about 1400 m above sea level) from years 1979 to 2014. The geographic area is from -66.25 to -20.25$^\circ$ E and from 5 to 30$^\circ$ N (Fig \ref{fig:sch_domain}). With a resolution of 0.25$^\circ$, each day is thus represented by a field of 101 x 189 values, transformed into a vector of 19,089 components. The data cover a period representing a base of 13,140 days.\\

\noindent The second dataset comes from the reanalysis of daily cumulative rainfall measured by satellite by the NASA's TRMM project, from 2000 to 2014, for the same geographic area and resolution as the first dataset. The data cover a period representing a base of 5,415 days. In order to assess the study area, surface rainfall data supplied by Meteo France (Guadeloupe and Martinique), from 1979 to 2014, are used in the design of the ED$_{RAINFALL}$. They allow to determine histograms edges as outlined in Section \ref{subsubsec:hist}.

\subsection{Issues generated by the L2 distance}
\label{subsec:prob_l2}
Most of previous studies in this domain \cite{article34,article5,Stephenson14,article36} use the same distance to compare two fields: the distance associated with the L2 norm. This distance between two vectors $V=(v_1,v_2,\ldots,v_n)$ of daily data for two days $d_1$ and $d_2$ is calculated by:

\begin{equation}
d_{L2}(V(d_1),V(d_2)) = \sqrt{\sum_{i=1}^n ( v_i(d_1)) - v_i(d_2) )^2}
\end{equation}

\noindent where $v_i(d_j)$ is the $i$-th value of the vector $V(d_j)$ of data for the day $d_j$ and $n$ is the number of daily data. Although L2 is commonly used in the clustering methods, we think that L2 is partly responsible for the difficulties encountered in computerized climate analysis study.

\begin{figure}[hbt!]
\centering
\includegraphics[scale=0.3]{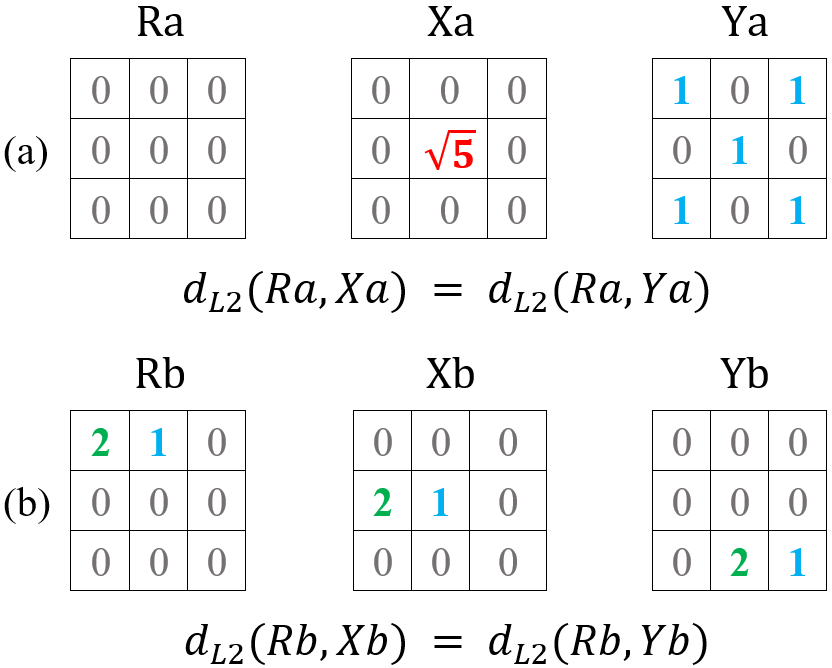}
\caption{Representation of the characteristics of the L2 distance: (a) a strong and localized fluctuation Xa produces the same L2 distance as a multitude of small variations Ya from the reference Ra, (b) whether a small spatial shift Xb, or a large one Yb, produces the same L2 distance from the reference Rb.}
\label{fig:sch_pb_l2}
\end{figure}

\noindent For the purpose of illustration, Figure \ref{fig:sch_pb_l2} shows two schematic examples of a reference field and two other situations. The distance L2 between the reference and each situation is the same, even though one is ``physically'' much closer to the reference that the other. Two reasons underlie such a behavior. First, when data are described in a large vector space, a multitude of small fluctuations that are spatially spread across the field can be considered to be as important as one single large and very localized fluctuation (Fig \ref{fig:sch_pb_l2}a). Second, a situation ($Xb$) that presents the same spatial structure but slightly translated from the reference ($Rb$) has the same L2 norm than a situation ($Yb$), where the spatial shift is large (Fig \ref{fig:sch_pb_l2}b). This is probably more sensitive when the clustering is applied to variables, such as rainfall, which are spatio-temporal intermittent fields. Two fields, slightly translated, do not have many pixels in common, even if they are similar. In the clustering process these side effects tend to skew the comparison between the daily spatial patterns and therefore affects the quality of the clusters.

\subsection{Design of the Expert Deviation}
\label{subsec:design_ed}
The effects of L2 in clustering methods have already been highlighted in many studies from other domains and other measures have been proposed \cite{article13,article48,article69,article37}. In this section, an original dissimilarity measure based on an Image Retrieval approach, is proposed. The conception of the measure consists in three phases: first, subdivide the area of interest into zones according to the specialists knowledge; second, build histograms for quantifying the variable of interest while reducing the influence of spatial location; third, apply a dissimilarity measure to the histograms. In the example chosen for this study, geographic segmentation into zones is performed according to atmospheric structures (Fig \ref{fig:sch_domain}).

\subsubsection{Partial management of spatialization}
\label{subsubsec:decoup}
For a number of computer vision applications, the image can be analyzed at the patch level rather than at the individual pixel level \cite{article65,article64,article63}. Image patches contain contextual information and have advantages in terms of computation and generalization \cite{article62}. From this point of view, the decomposition of an image into patches or zones that do not overlap, provides a simple but effective way of overcoming the curse of dimensionality \cite{article70}. In this work, we subdivided the cumulative rainfall and wind fields of the North Atlantic tropical area into four zones (Fig \ref{fig:sch_domain}), to limit the computation time and demonstrate the value of the approach without seeking to optimize it. These zones are defined in order to take the knowledge of experts into consideration. Thus, three zones correspond to specific and known centers of action. In Fig \ref{fig:sch_domain}, A1 is the western depressions forming zone; A2 is the North Atlantic Subtropical High (NASH) zone; the zone of interest, called A3, includes the landmass, i.e. the continental zone and the arc of the Lesser Antilles; and A4 is the zone of low pressure linked to the Intertropical Convergence Zone (ITCZ). To compare two fields of values, we subdivided the fields according to these four zones and then compared them with each other.

\subsubsection{Comparison of intensity distribution}
\label{subsubsec:hist}
Once each field has been spatially subdivided, we no longer have to pinpoint the exact location of the phenomena in each zone. It seems relatively reasonable to ignore their position down to the exact mesh, and instead to look at the distribution of each fields of the datasets, ignoring in this way the notion of spatial location. We have opted for a discrete representation of the data by frequency histograms, in order to estimate the distribution of the intensities. An estimation of continuous probability densities would have raised unecessary parameterization problems. Moreover, quantification might help reducing the effects of the small fluctuations reported in Section \ref{subsec:prob_l2}, even though edge effects around the boundaries of the selected intensity classes would remain. For the wind dataset, the Beaufort scale is used to set histogram edges for representing wind speed distribution. For computing the rainfall dataset histograms, bins are determined from the rainfall data collected in the area (Table \ref{tab:bornes}). We selected eight bins of possible intensities. The boundaries of these bins were selected so that the distribution of the rainfalls over these bins is uniform.

\begin{table*}[hbt]
\centering
\caption{Boundaries of the histogram classes used to quantify daily rainfall data. These edges are determined from rainfall records of the study area.}
\small
\begin{tabular}{|rccccccccc|}
\hline
  \textbf{Centiles ($\%$)} & 0 & 0.35 & 0.5 & 0.7 & 0.8 & 0.9 & 0.95 & 0.99 & 1 \\
\hline
  \textbf{Rainfall ($mm$)} & 0 & ]0,1.2] & ]1.2,2.2] & ]2.2,5.2] & ]5.2,8.7] & ]8.7,16.4] & ]16.4,26.9] & ]26.9,59.2] & ]59.2,+$\infty$[ \\
\hline
\end{tabular}
\label{tab:bornes}
\end{table*}

\begin{figure}[hbt!]
\centering
\includegraphics[scale=0.26]{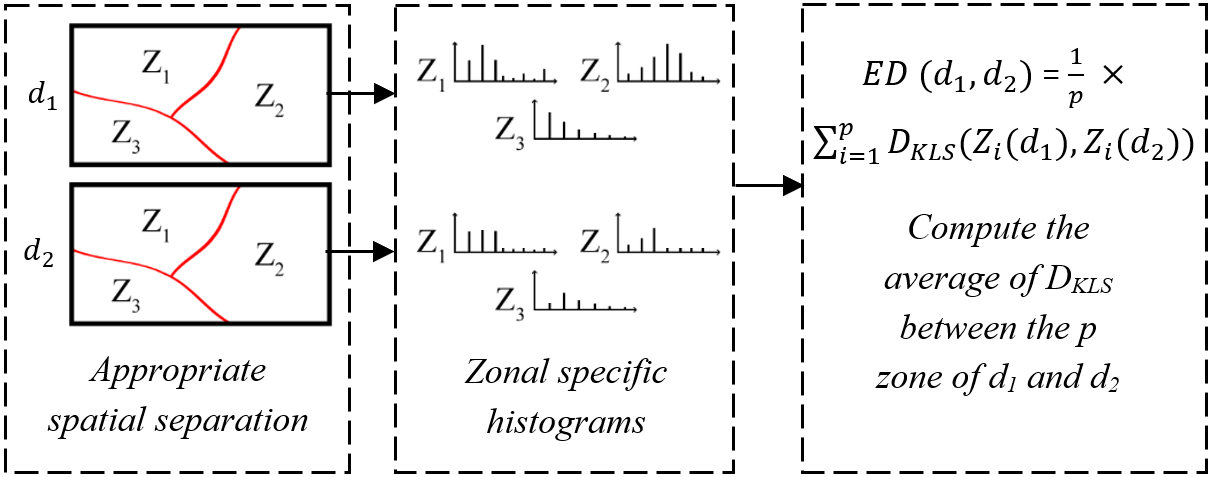}
\caption{Schema showing the computation process of the expert deviation: zonal quantification using custom edges, the use of symmetrized Kullback-Leibler divergence ($D_{KLS}$) on each zone to get four values and the computation of the average to obtain $ED (d_1, d_2)$}
\label{fig:sch_design_ed}
\end{figure}

\noindent For comparing histograms without defining a specific distribution, the Kullback-Leibler symmetrized divergence appears to be a judicious choice \cite{article64,article63,article62}, with a formula as follows:
\begin{equation}
\begin{array}{rl}
D_{KL}(P,Q)& = \sum_{c=1}^m D_{KL}(P(c),Q(c))\\[0.5cm]
 & = \sum_{c=1}^m P(c) log\frac{P(c)}{Q(c)}\\[0.5cm]
D_{KLS}(P,Q)& = D_{KL}(P,Q) + D_{KL}(Q,P) \\
\end{array}
\end{equation}

\noindent where $P$ and $Q$ are two distributions of discrete probabilities, $c$ is the index of a bin for each distribution and $m$ the number of bins. Although it is possible to use the Kullback-Leibler symmetrized divergence with continuous fields, we thought it was more interesting to quantify the data. The distinct intensity distributions obtained are then used to compute the Kullback-Leibler divergence in each zone. The average of the divergences by zone provides the dissimilarity between two days. We named Expert Deviation, referred to as ED, the quantity defined by:

\begin{equation}\label{eq:ed}
	ED(d_1,d_2) = \frac{1}{p} \times \sum_{i=1}^p D_{KLS}(Z_i(d_1),Z_i(d_2))
\end{equation}

\noindent where $d_1$ and $d_2$ are two days, $Z_i(d_j)$ is the histogram of the zone with reference $i$ and $p$ the number of zones ($p=4$ in this study). All of the operations listed above are summarized in Fig \ref{fig:sch_design_ed}, a specific ED being designed for including expertise concerning each dataset (e.g. Beaufort scale for the wind).

\subsubsection{Clustering assessment}
\label{subsubsec:sil}
In order to assess and compare the results produced by the ED method with those produced by L2, we use the silhouette index which allows to quantify the internal homogeneity of each cluster and the separation between the different clusters \cite{article29,Boltz2007}. This index also provides information in the selection of the number of clusters and the algorithm to be retained from the clustering analysis. This step was not always carried out in the previously mentioned studies, misconsidering the visual coherency of the centroid patterns as being sufficient to validate the clustering. Of course, the computation of the silhouette index also integrates ED to perform an effective evaluation of the results from clustering using ED.

\subsubsection{Integration in clustering analysis}
\label{subsubsec:clus_int}
The integration of these new dissimilarity measure in clustering algorithms to replace L2 is the next step. By default, in KMS, at the end of each iteration, the centroids of each cluster are calculated as their barycenter. These centroids are the average field computed over each cluster. We think that these centroids are artificial and not representative of realistic spatial configuration since the average smooths every small spatial structure, resulting in huge structure that appears in no observed field. Hence, the clustering process leads to group together the days around a non relevant field. This method is not efficient for intermittent fields such as cumulative rainfall. Therefore, we propose firstly to define the centroid as a group of four histograms which correspond to the geographic zones previously introduced in Section \ref{subsubsec:decoup}. Secondly, in their visual analysis process of the clusters, other authors rely once more on these centroids as representative of each cluster, raising the same questions about their relevance. In order to get a significative (or, at least, a more relevant) view of each cluster center, we propose to select the nearest element from the centroid according to ED. This center is now an existing field.

\section{Results}
\label{sec:results}
\subsection{Numerical assessment of the quality of the clusters}
\label{subsec:res_num}
\begin{figure}[hbt!]
\centering
(a)\includegraphics[scale=0.49]{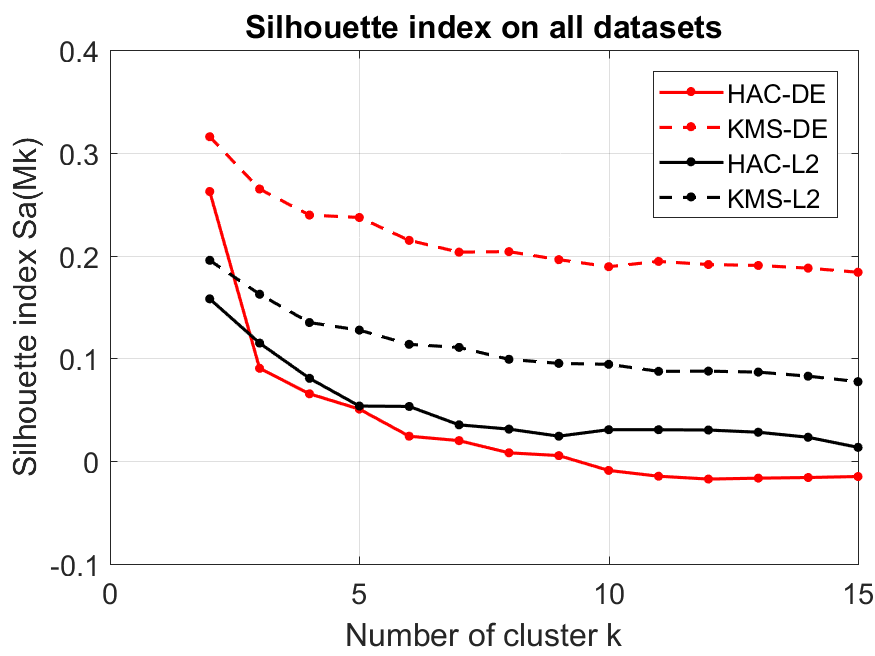}
(b)\includegraphics[scale=0.28]{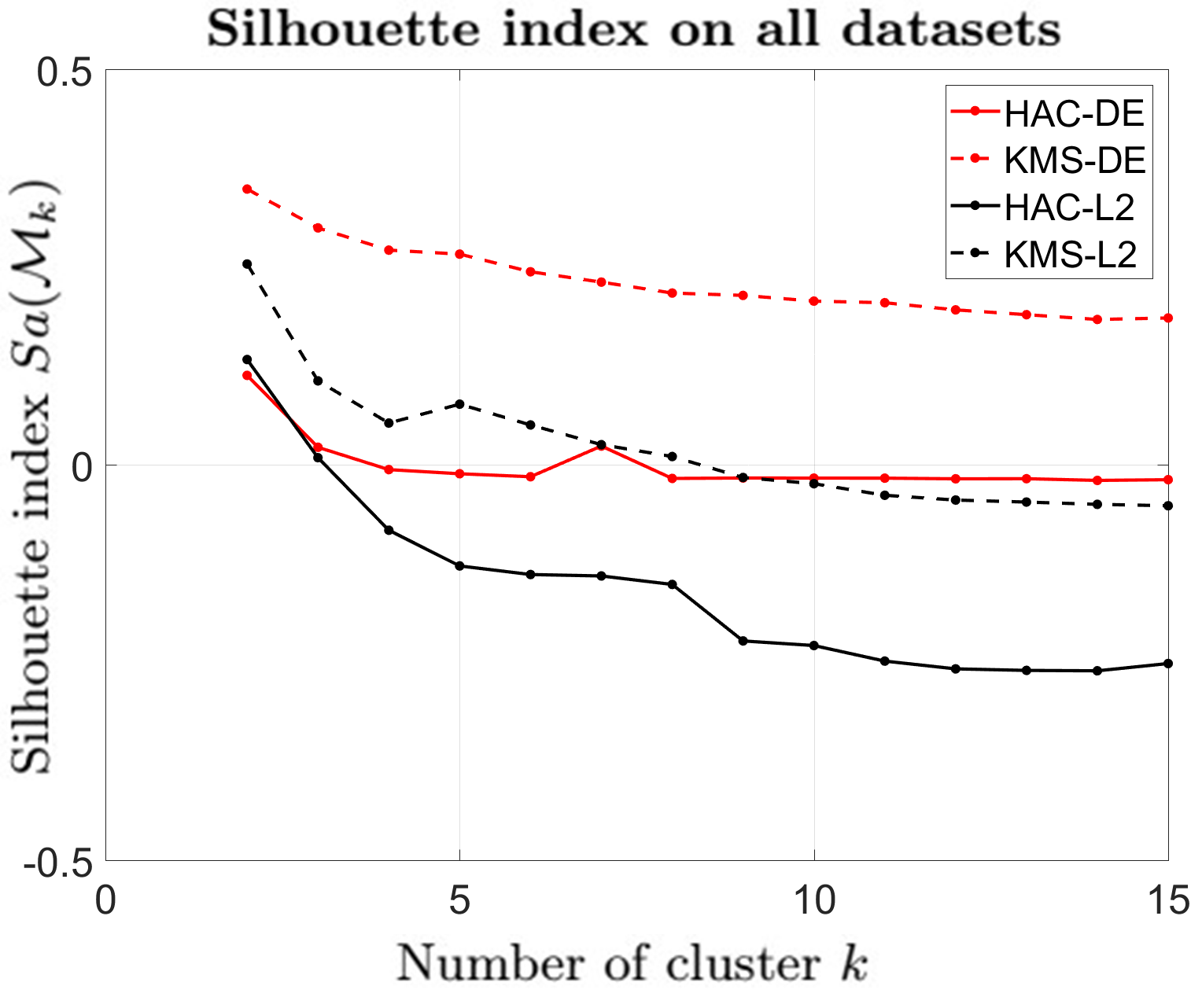}
\caption{Silhouette index evolution n function of k, the number of clusters - HAC (solid line), KMS (broken line), using L2 (black), using ED (red). Results for clustering of wind dataset in (a) and for rainfall dataset in (b).}
\label{fig:sch_silhouette}
\end{figure}

To reveal the possibilities of using these two measures, the numerical assessment of the quality of the clusters obtained by HAC-L2, KMS-L2, HAC-ED and KMS-ED was performed by computing the silhouette coefficients. Figure \ref{fig:sch_silhouette} (a) and (b) show the evolution of the silhouette coefficient as a function of the number k of clusters for both datasets. It reflects the combined evaluation of the proximity of an element to the elements of its cluster and the distance of the cluster to all of the other elements. This result highlights two interesting points as follow:\\

\noindent 1) HAC algorithms produce coefficients that are notably lower than those obtained by KMS type algorithms. For HAC-ED$_{WIND}$, the silhouette coefficient is irrelevant. HAC-L2$_{RAINFALL}$ silhouette coefficient is quite better; Regarding HAC-ED$_{RAINFALL}$, the silhouette coefficient nears zero as soon as the number of searched clusters (k) is greater than three, indicating the irrelevance of the possible clusters detected. Worse, in the case of HAC-L2$_{RAINFALL}$, the silhouette coefficient becomes negative for the same value of k, indicating that many points are assigned to clusters that do not represent the best possible choice.\\

\noindent 2) The numerical values obtained for KMS-L2 and KMS-ED are more instructive than the results obtained by HAC. Although the shape of the curves is generally decreasing with k, we can observe a slight inflection around k = 5 for both datasets. For this value of k, KMS-L2$_{RAINFALL}$ has a silhouette value of 0.08, while KMS-ED$_{RAINFALL}$ has a value of 0.26. KMS-L2$_{WIND}$ has a silhouette value of 0.13, while KMS-ED$_{WIND}$ has a value of 0.23. As before, the results obtained by the method based on L2 are very lower to those obtained by the method based on ED. In addition, only the KMS-ED$_{RAINFALL}$ method obtains silhouette coefficients greater than 0.20 for a significant number of clusters and KMS-ED$_{WIND}$ produce the best results for wind dataset, thus indicating the presence of relevant structures, although weakly marked within the datasets.\\

\noindent Numerically, the introduction of the Expert Deviation enabled us to obtain clusters that seem relevant (silhouette~$>$~0.2) whereas L2 gives clusters without any coherence (silhouette $\simeq$ 0.1). The low silhouette index of the methods based on L2 probably explain that previous studies did not present such numerical evaluation.

\subsection{Spatio-temporal dynamics analysis}
\label{subsec:res_dynm}
In order to strengthen the analysis of this comparative study, the results obtained by both methods were reviewed by atmospheric physicists. This sub-section presents their observations.

\begin{figure*}[hbt!]
\centering
(a)\includegraphics[scale=0.06]{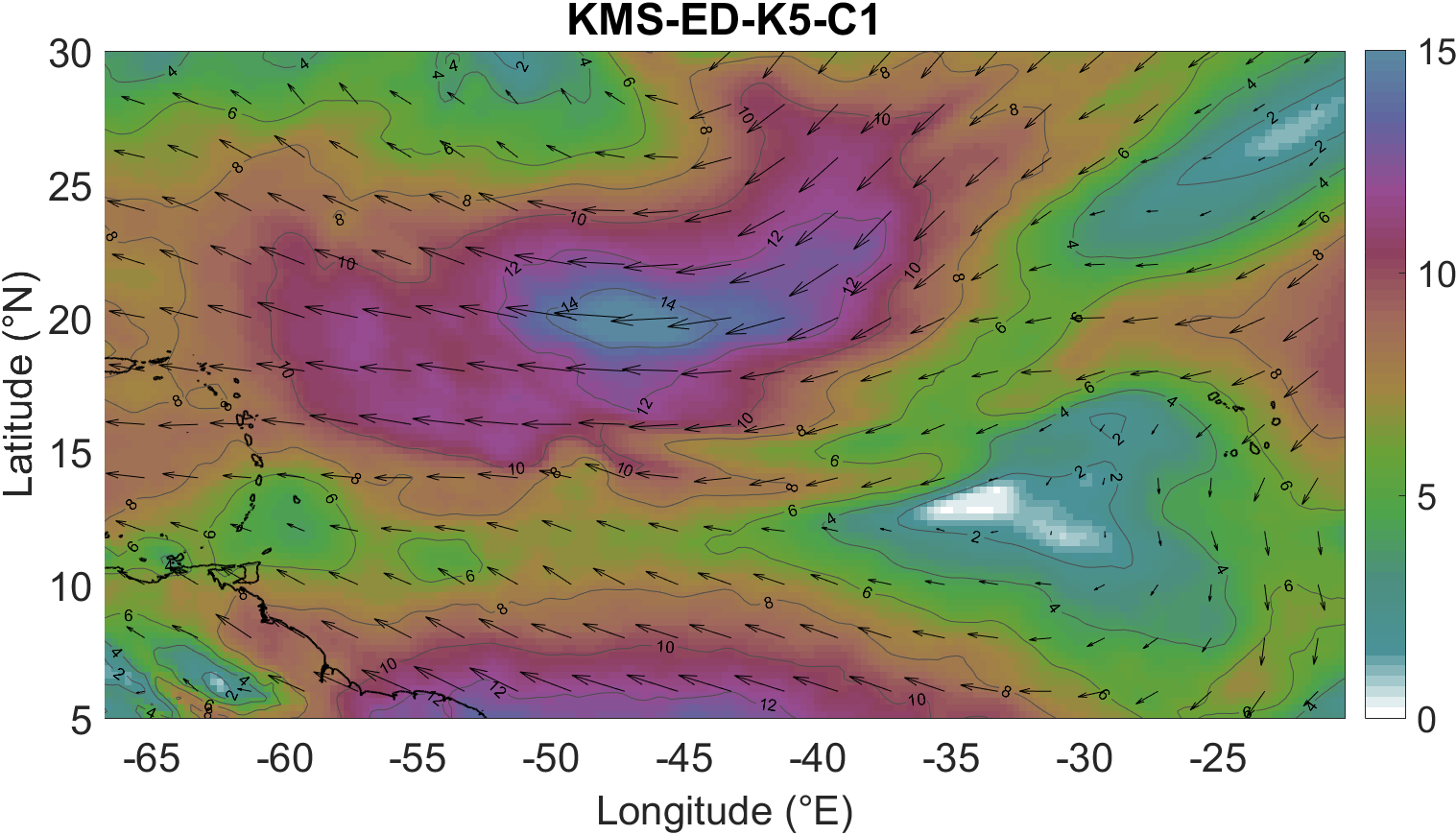}
\includegraphics[scale=0.06]{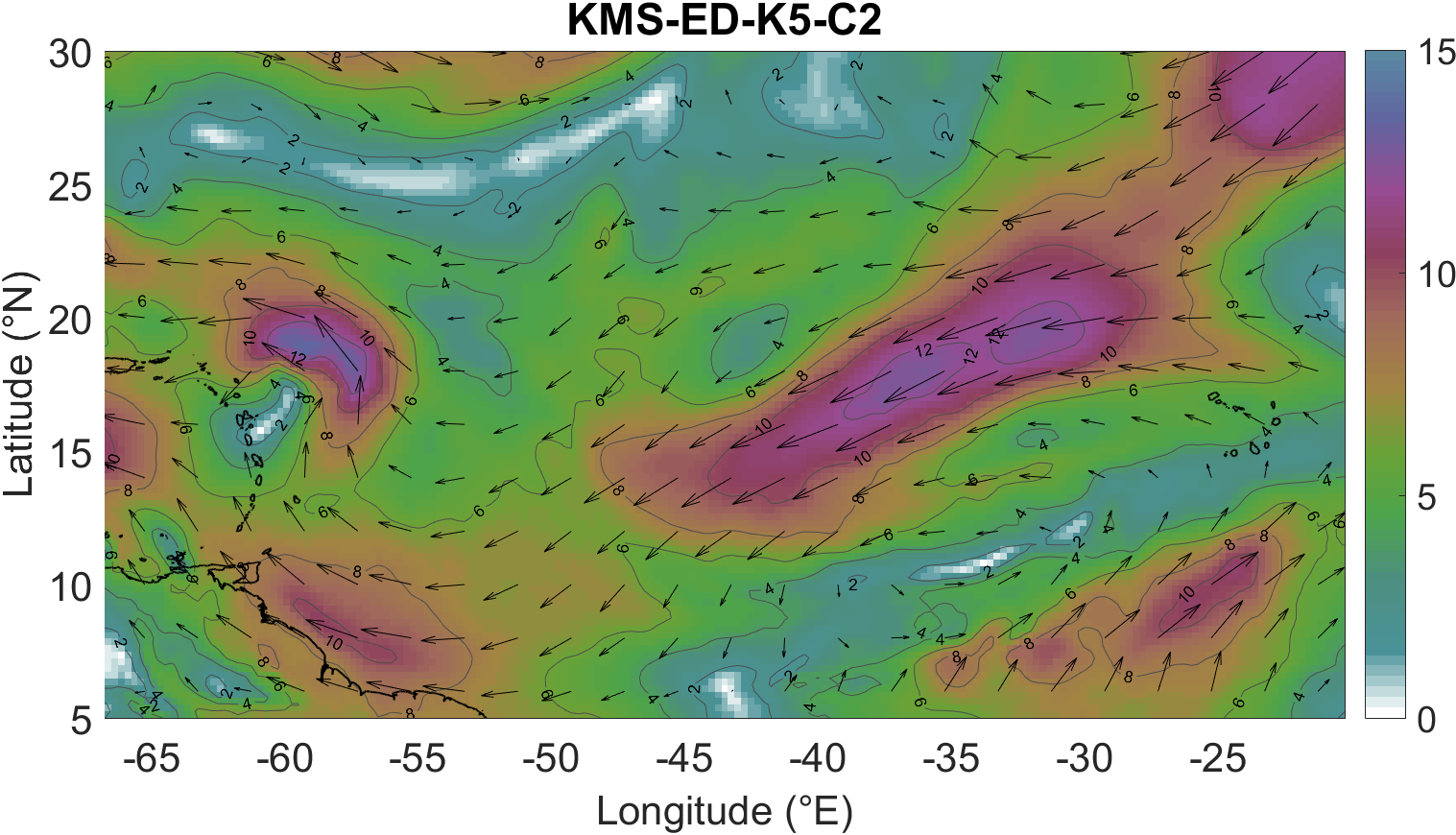}
\includegraphics[scale=0.06]{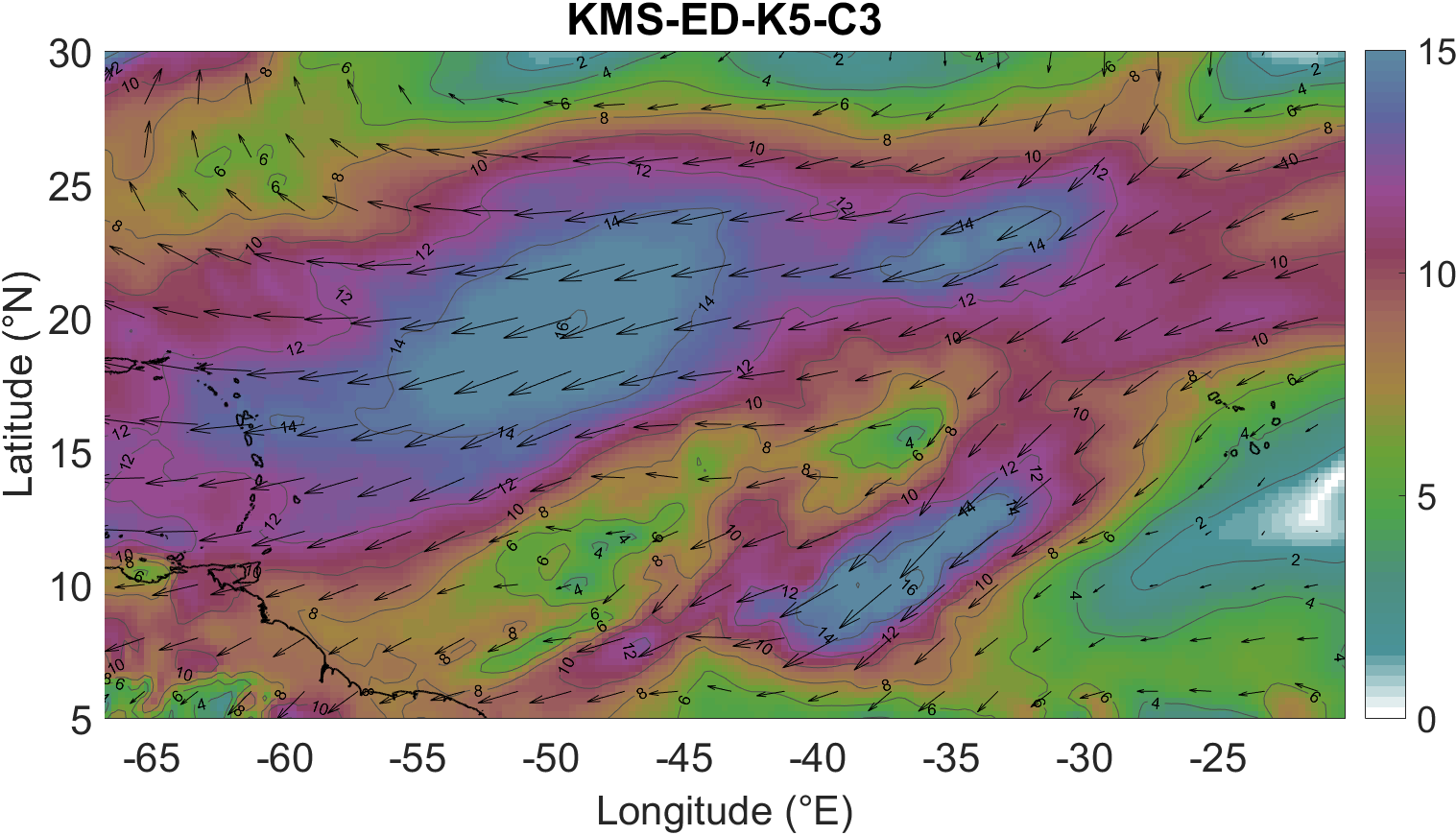}
\includegraphics[scale=0.06]{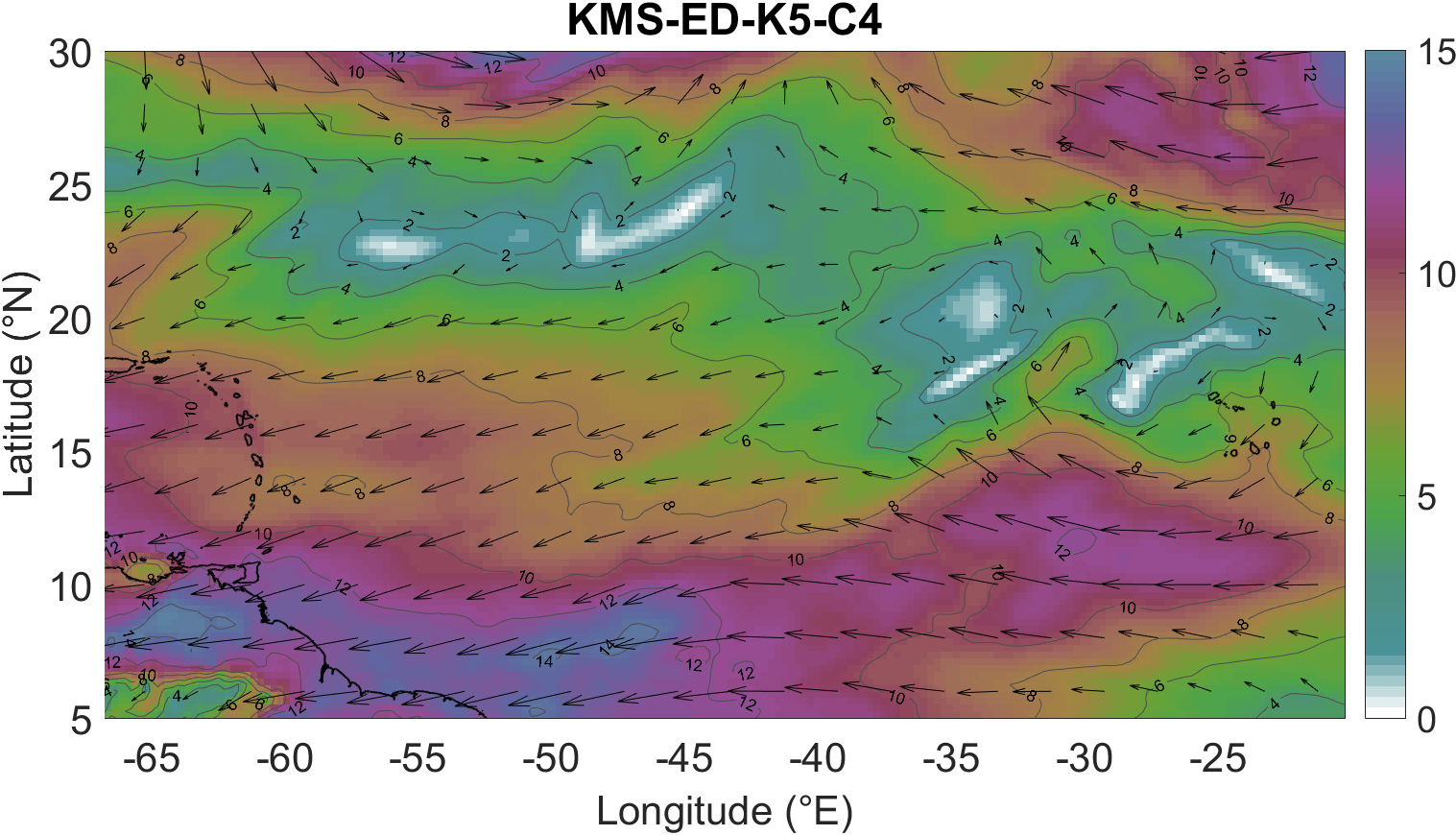}
\includegraphics[scale=0.06]{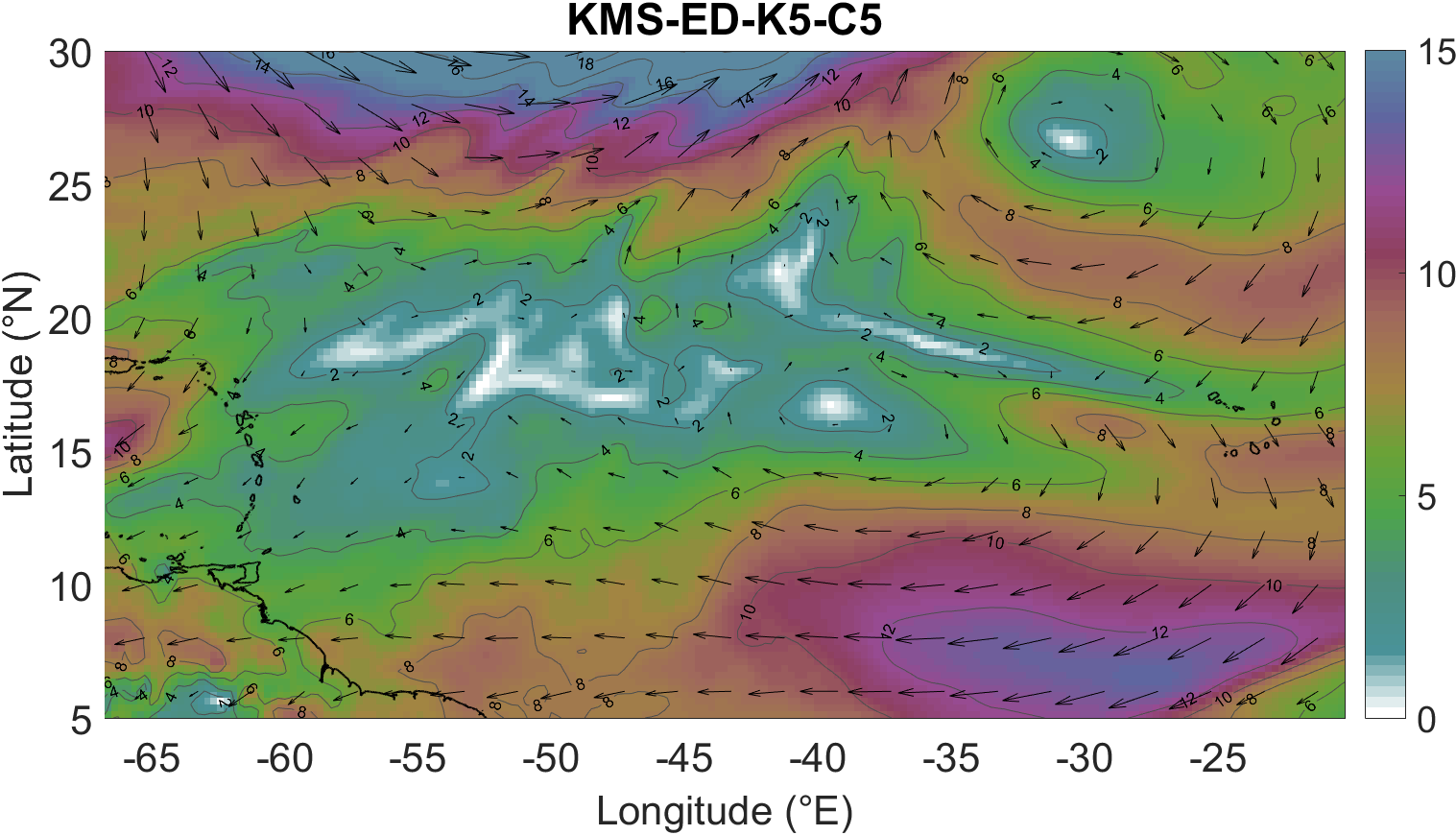}\\
(b)\includegraphics[scale=0.06]{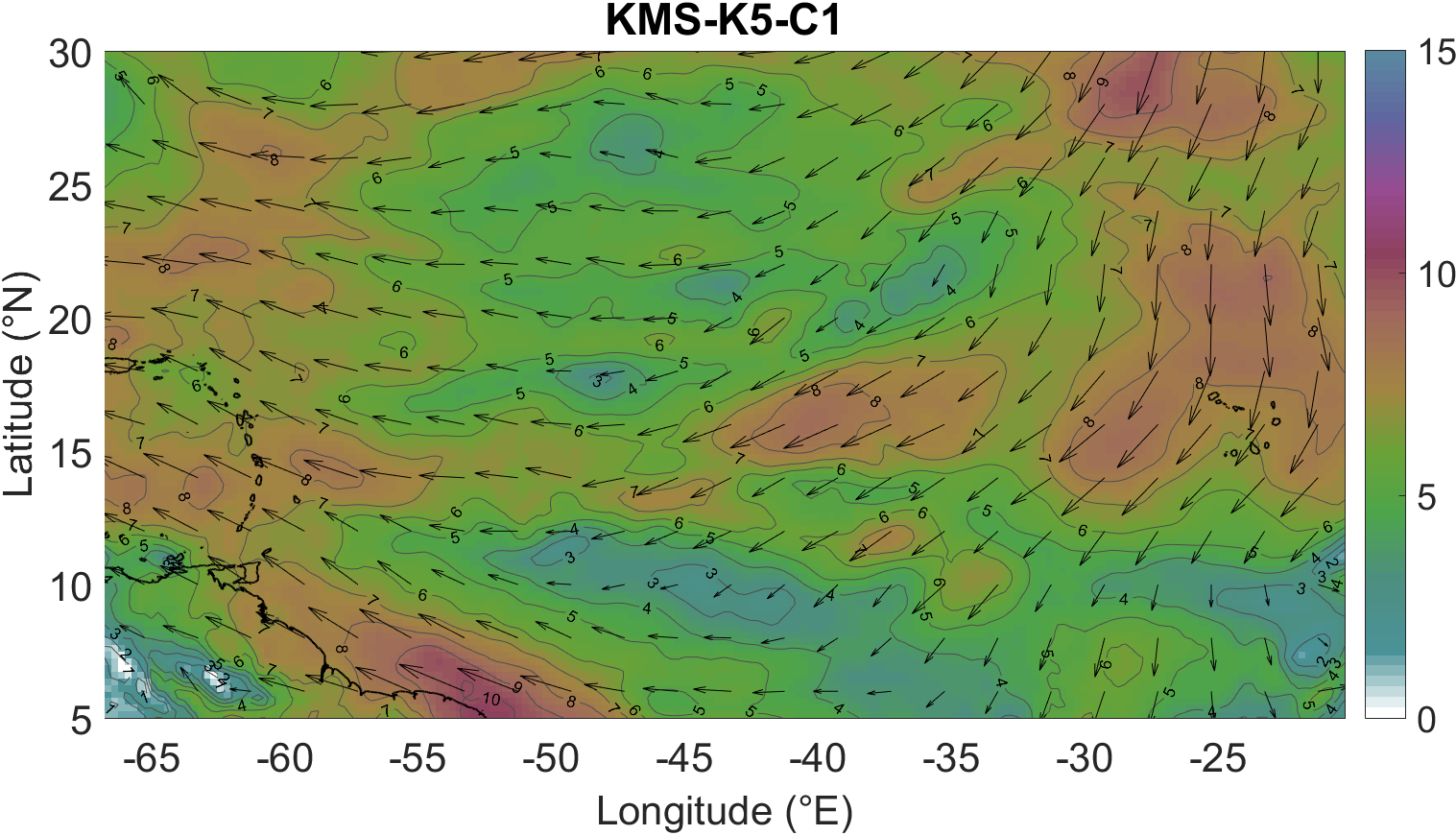}
\includegraphics[scale=0.06]{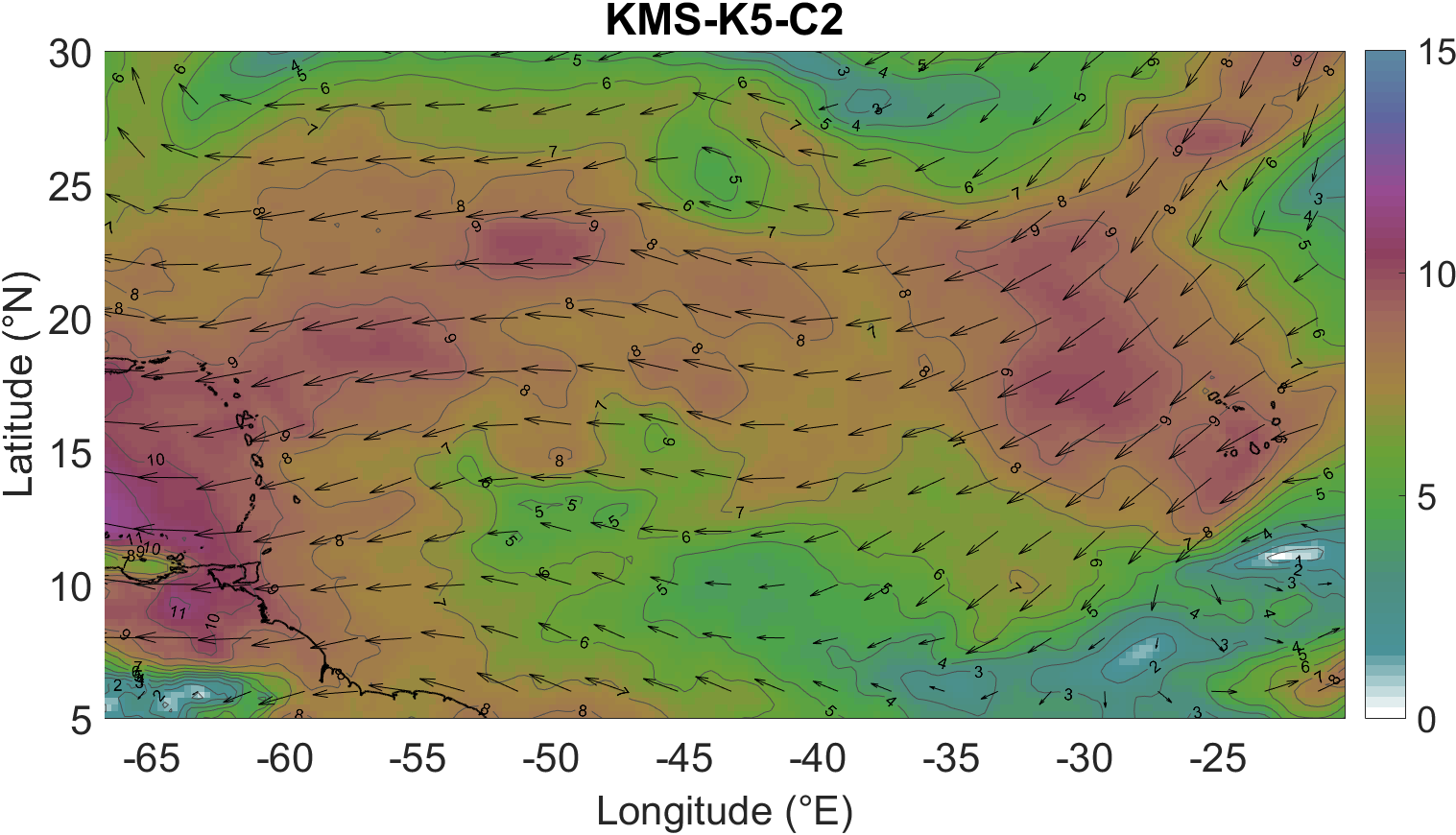}
\includegraphics[scale=0.06]{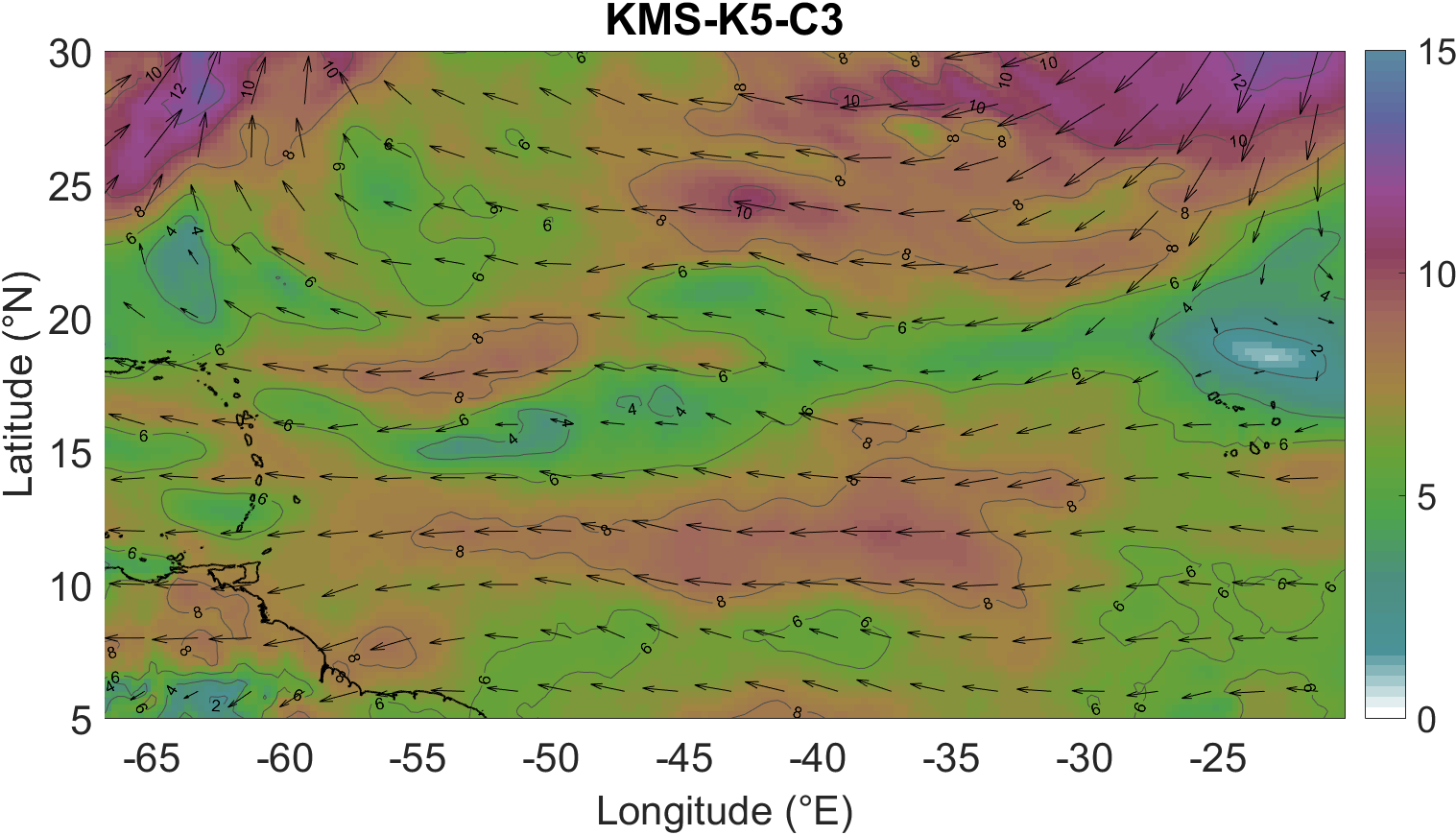}
\includegraphics[scale=0.06]{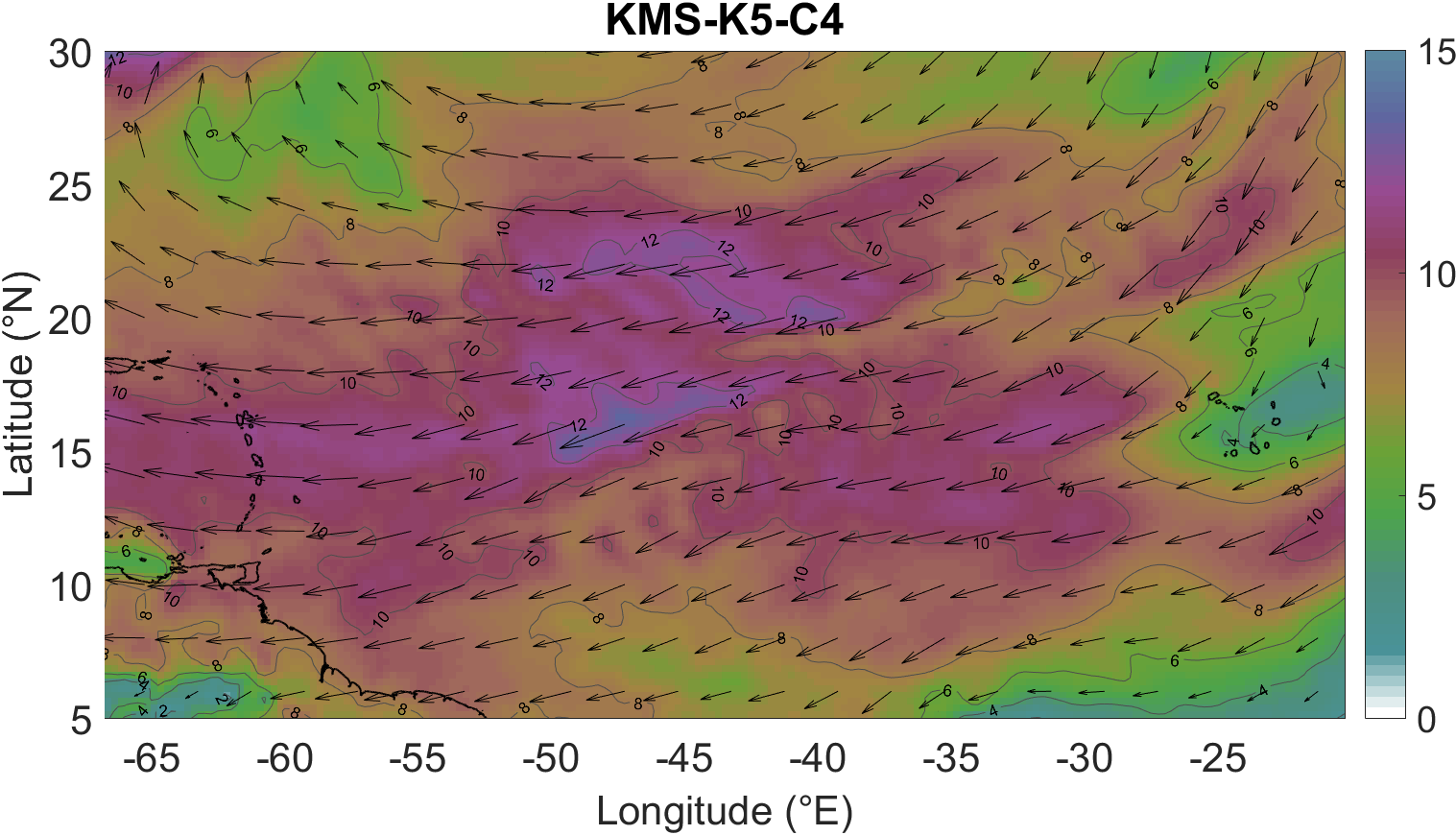}
\includegraphics[scale=0.06]{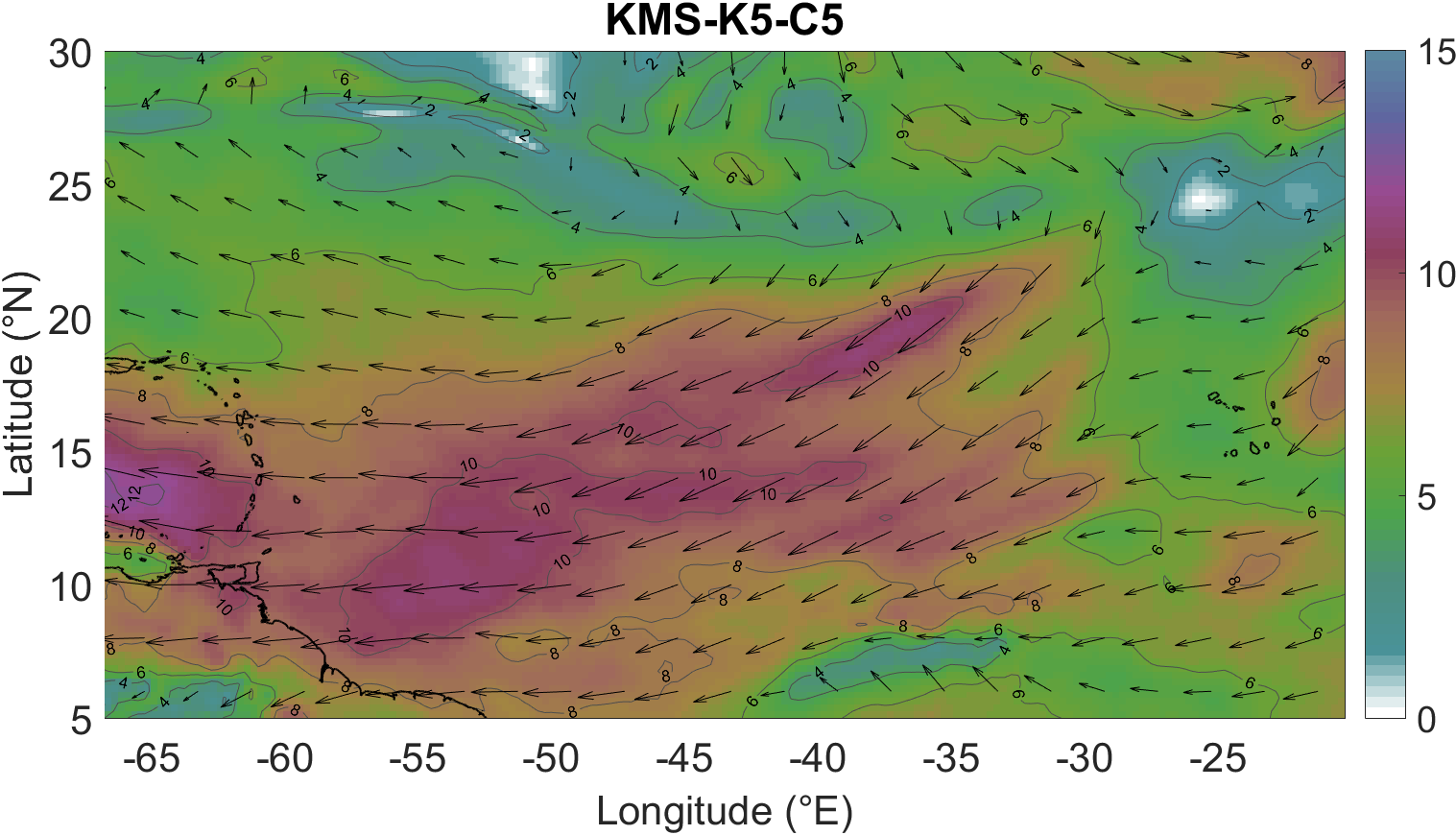}
\caption{Comparative graph of representative elements of the clusters of wind at 850 hPa from the KMS-ED$_{WIND}$ (a) and KMS-L2$_{WIND}$ (b) method, with $k = 5$.}
\label{fig:sch_wind_clus}
\end{figure*}

\begin{figure*}[hbt!]
\centering
(a)\includegraphics[scale=0.13]{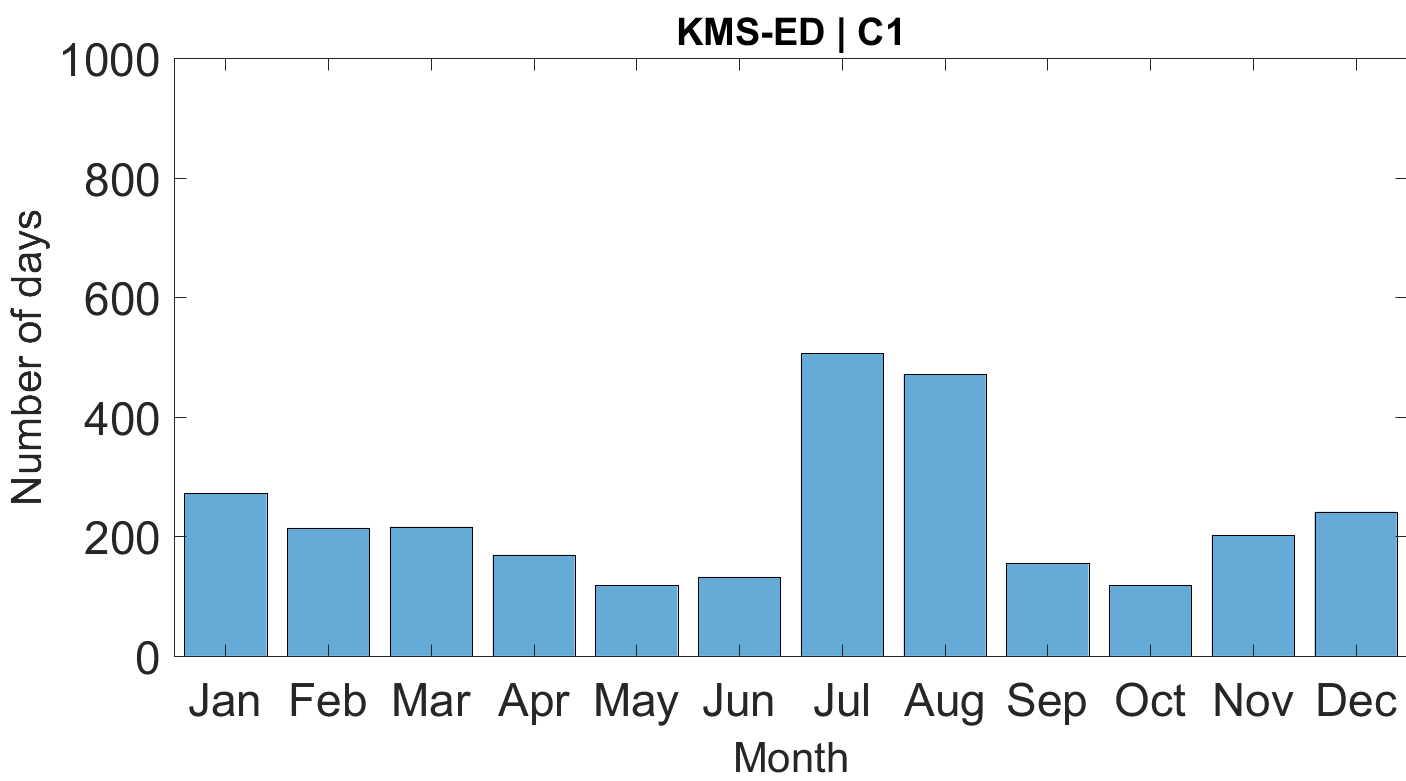}
\includegraphics[scale=0.13]{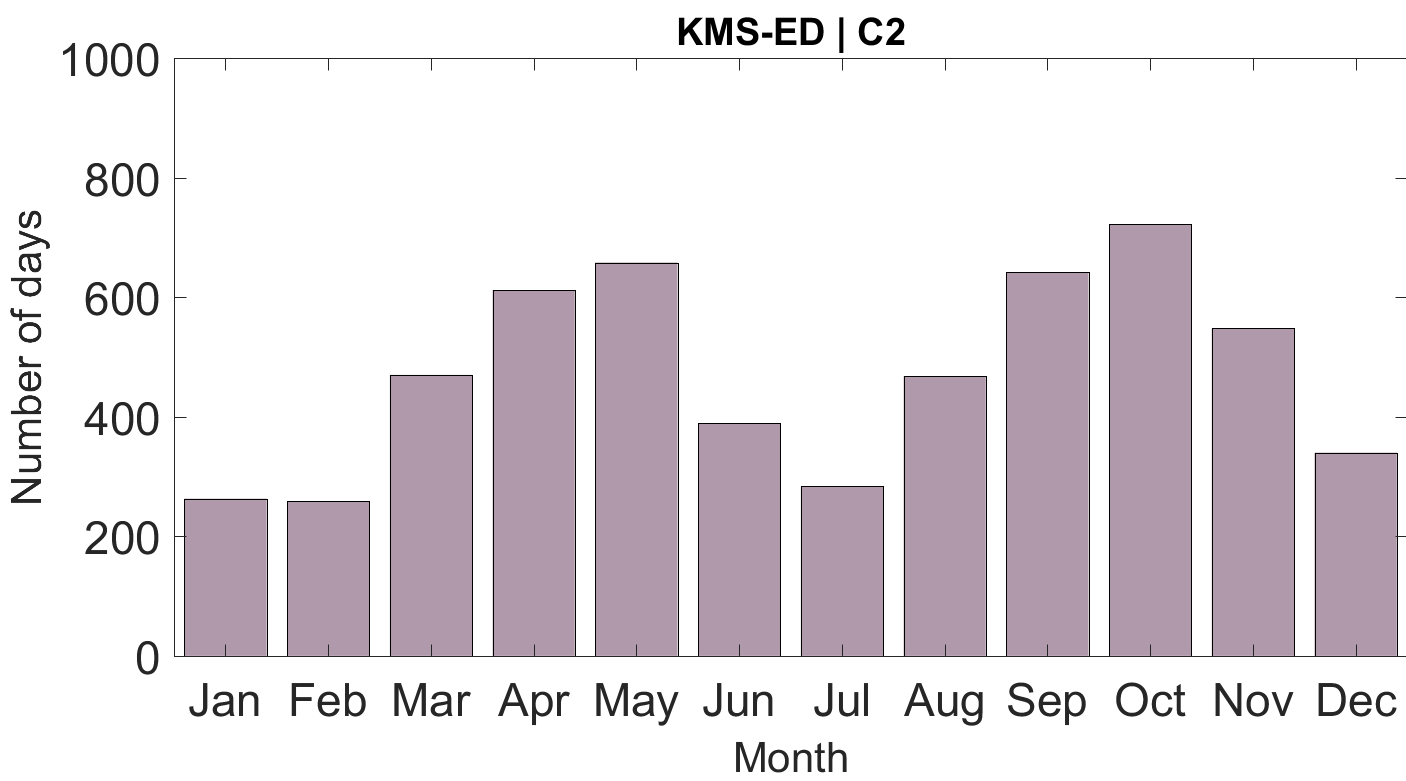}
\includegraphics[scale=0.13]{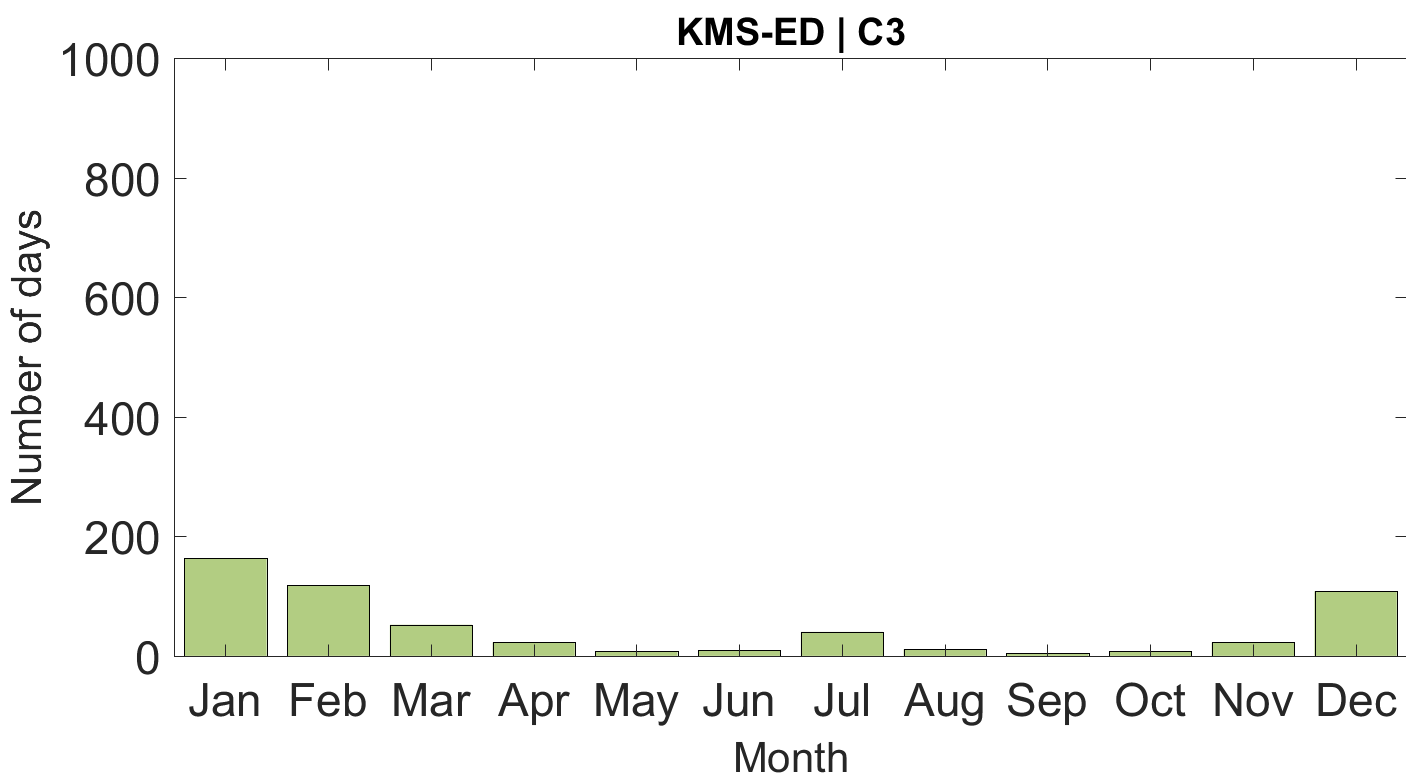}
\includegraphics[scale=0.13]{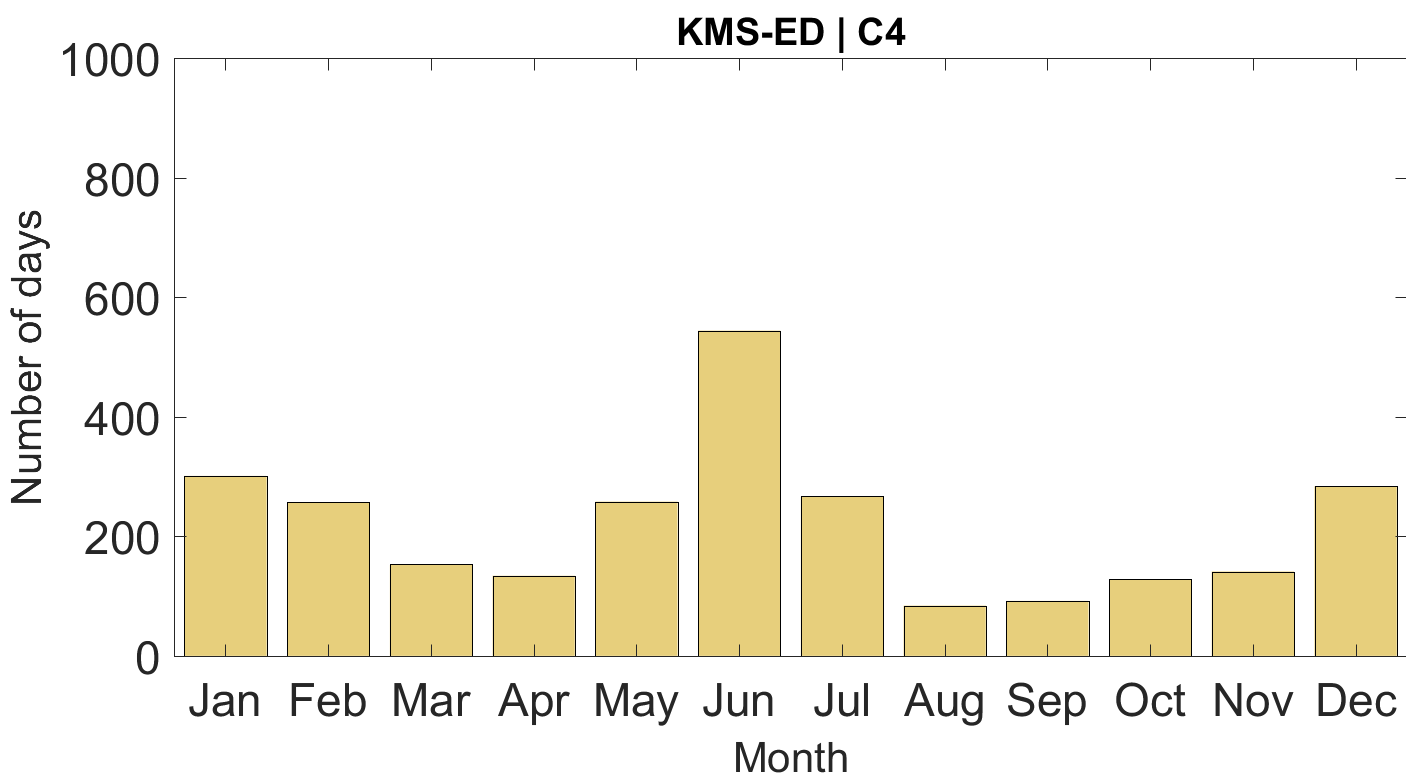}
\includegraphics[scale=0.13]{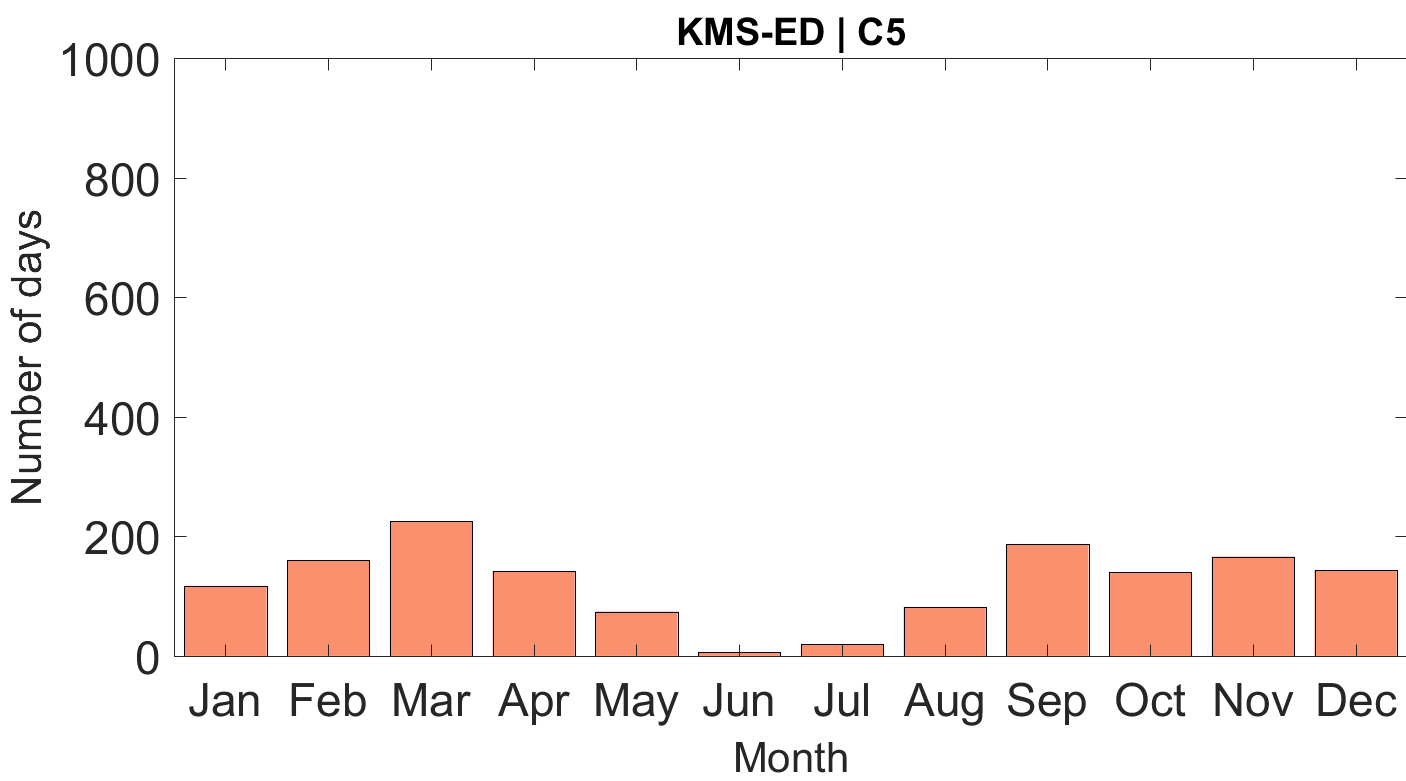}\\
(b)\includegraphics[scale=0.13]{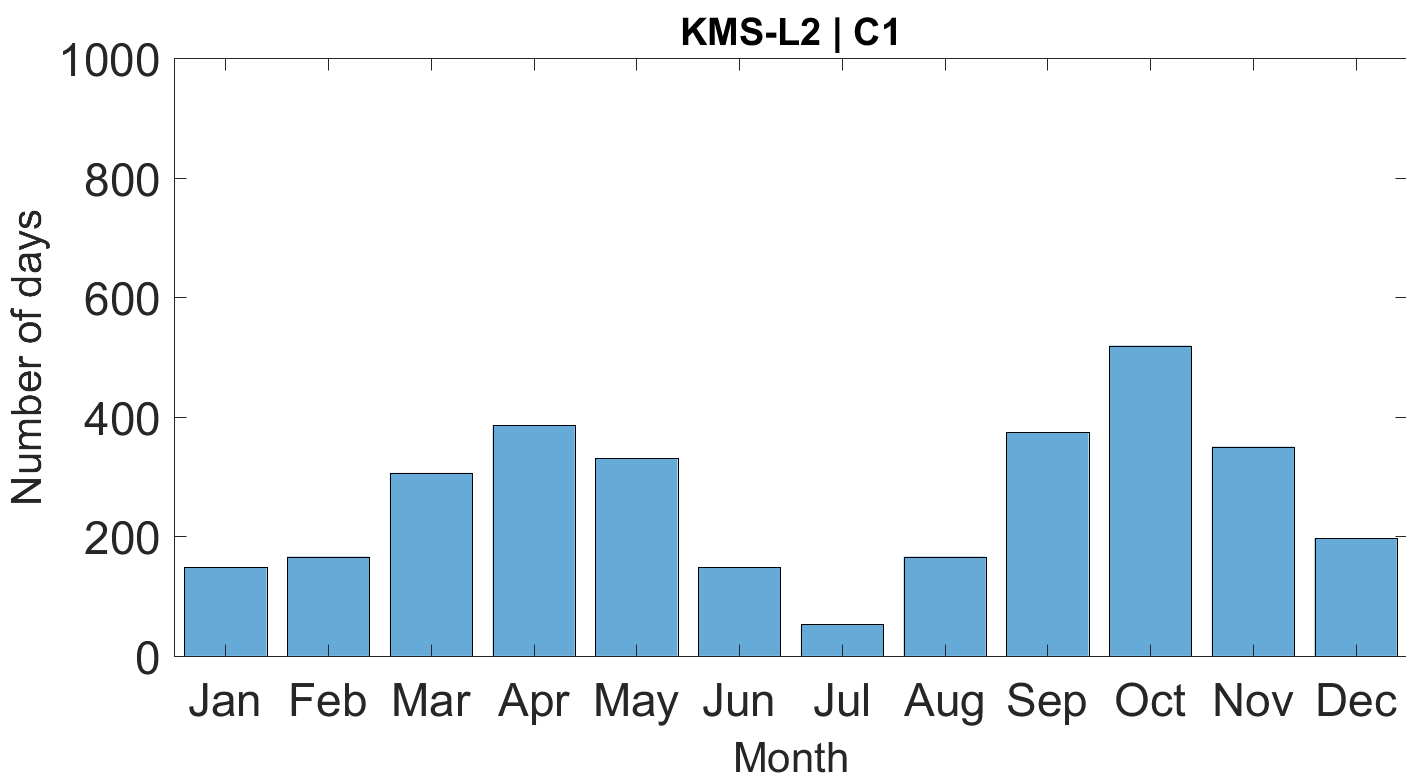}
\includegraphics[scale=0.13]{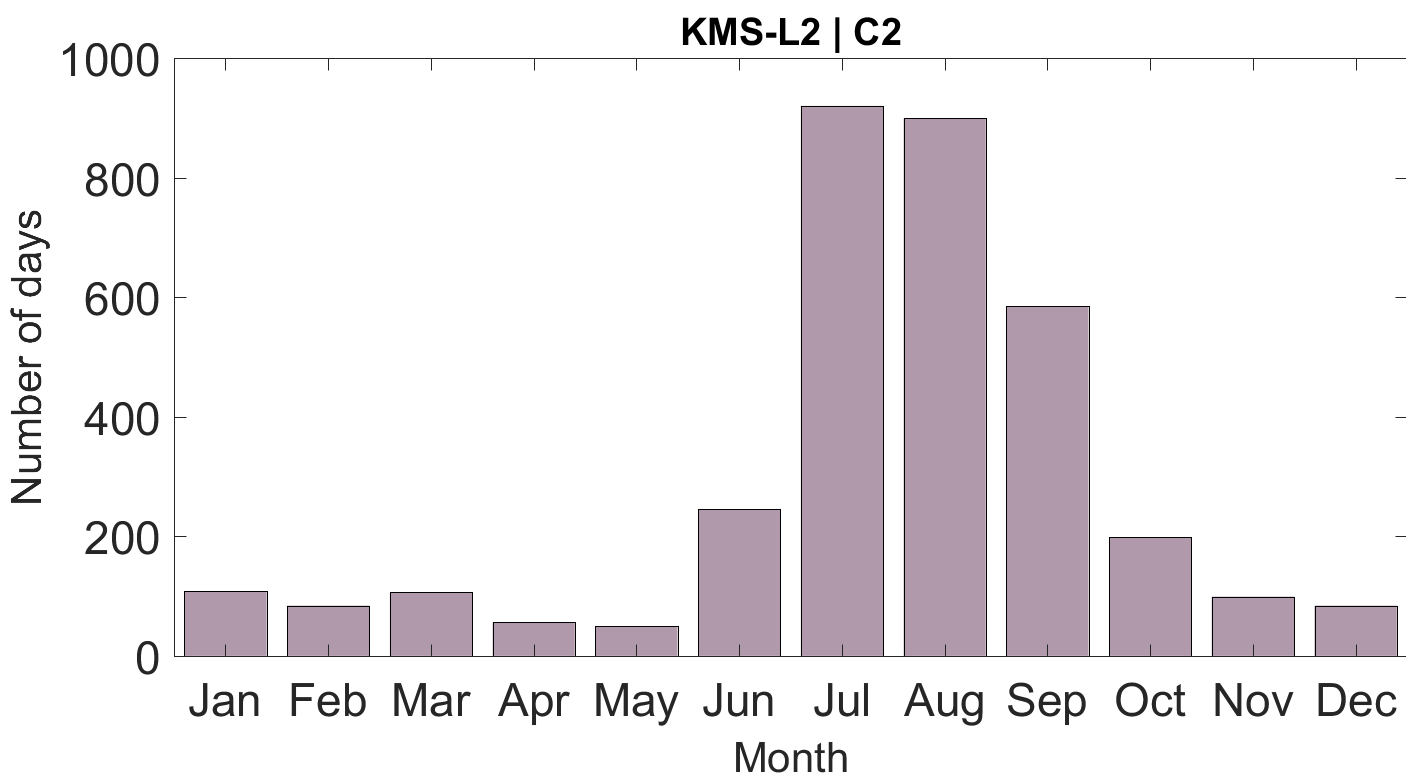}
\includegraphics[scale=0.13]{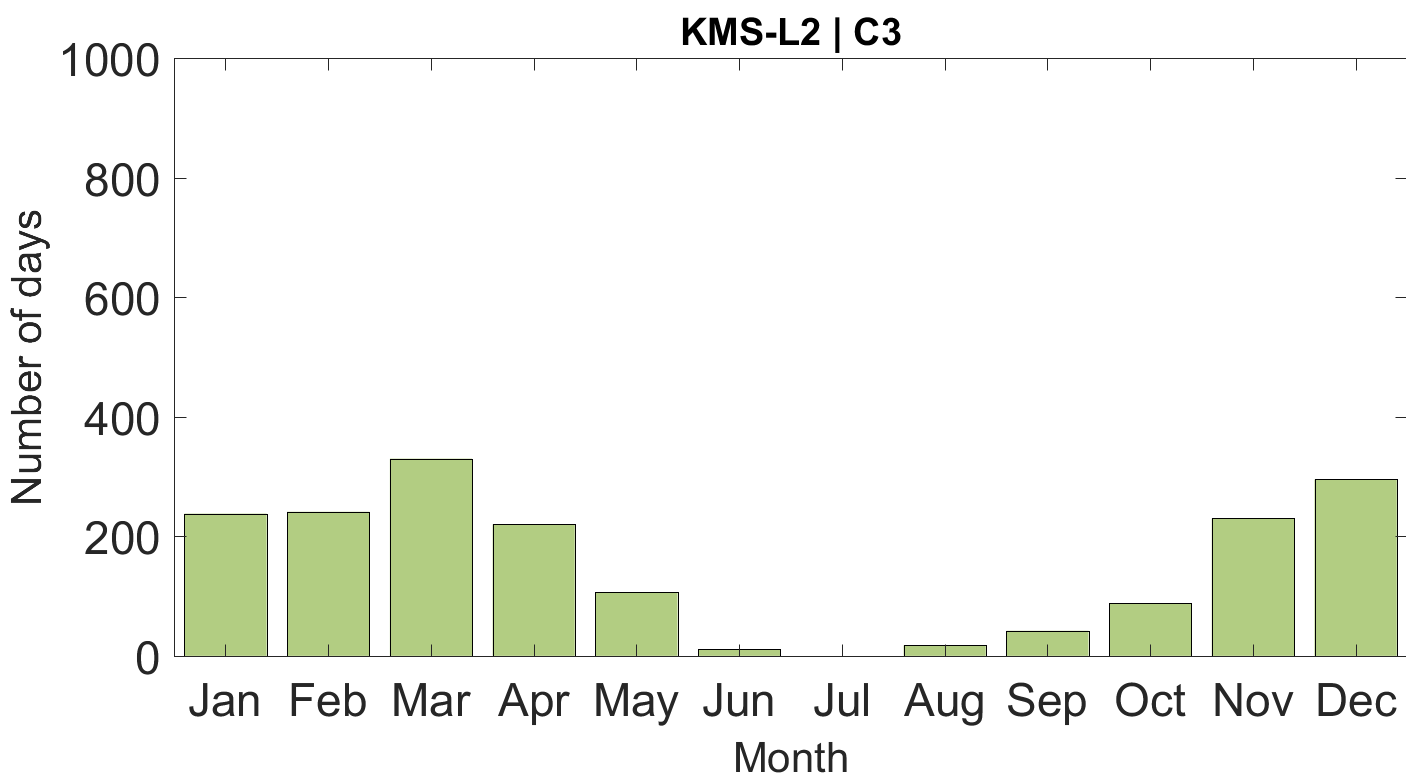}
\includegraphics[scale=0.13]{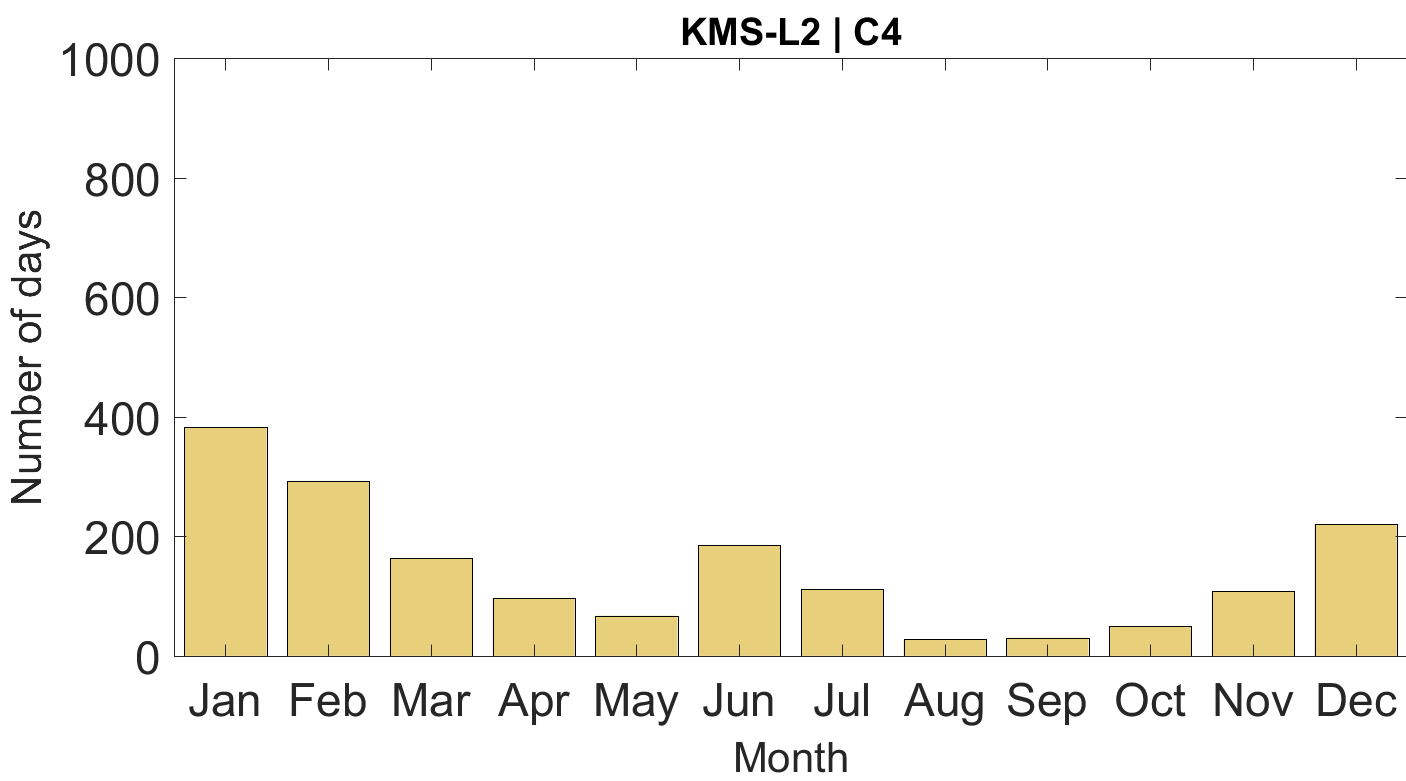}
\includegraphics[scale=0.13]{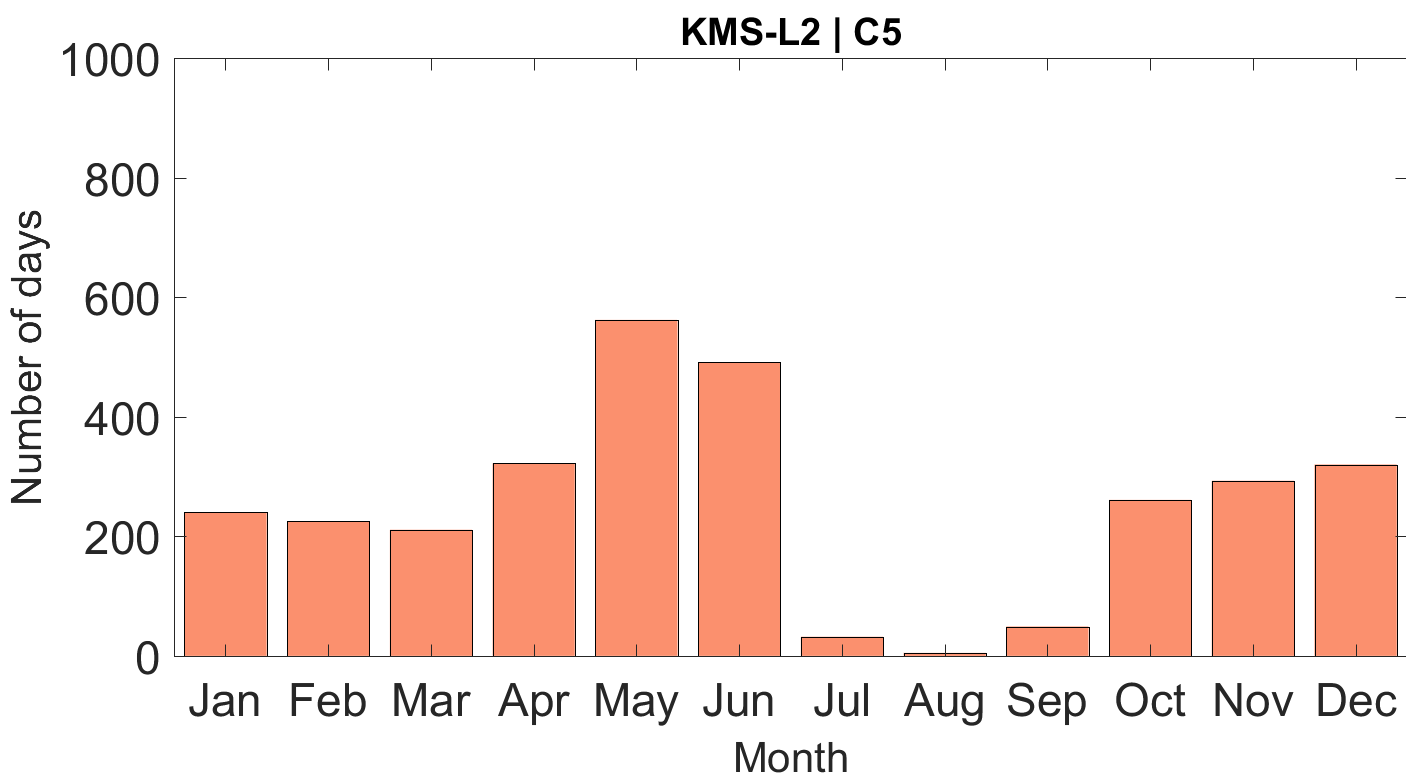}
 \caption{Monthly distribution of clusters using the KMS-ED$_{WIND}$ (a) and KMS-L2$_{WIND}$ (b) method over the year for the period 1979 to 2014.}
\label{fig:sch_wind_dist}
\end{figure*}

\noindent In Figure \ref{fig:sch_wind_clus}, to enhance the analysis, we present the representative elements of each cluster found with KMS-ED$_{WIND}$ and KMS-L2$_{WIND}$. According to Figure \ref{fig:sch_wind_clus}b, three of the five parterns of KMS-L2$_{WIND}$ show an increase in wind intensity over southern Lesser Antilles. Furthermore the five situations represented are spatially quite smooth, the usual atmospheric structures (such as NASH or ITZC) are not clearly visible. KMS-L2$_{WIND}$ does not produce clusters with a distinct monthly dynamic. Moreover there is a strong similarity in distribution shape of clusters C3, C4 and C5 (Fig \ref{fig:sch_wind_dist}b). 

\noindent For KMS-ED$_{WIND}$ the effect of patches can be observed (Fig \ref{fig:sch_wind_clus}a). Indeed, the configurations are all different and the usual atmospheric structures are clearly identifiable. It is now possible to interpret these situations for experts. In KMD-ED$_{WIND}$ C1, the trade winds are south-westerly and of low intensity spread over a narrow maritime strip that spreads to the south of the Lesser Antilles. One can also see a divergence phenomenon in atmospheric circulation, leading to stronger trade winds, especially in the northern part of the Lesser Antilles. C2 shows several divergence bands which produce very localized strong winds under the North Atlantic Subtropical High (NASH) which is located westward between 25 and 30$^\circ$N. The configuration promotes a weakening of the trade winds with south-easterly component as they approach the arc of the Lesser Antilles. It could indicate the tropical easterly wave passages over the Lesser Antilles. In C3, trade winds are strong and have an east to northeast component. C4 presents the NASH giving increasingly strong winds from the north of the Lesser Antilles to the South American continent. C5 show a situation of trade wind failure in the Lesser Antilles.

\noindent Figure \ref{fig:sch_wind_dist}a gives an overview of the monthly repartition of clusters. KMS-ED$_{WIND}$ produces clusters with a distinct monthly dynamic specific to each cluster.

\begin{figure*}
\centering
(a)\includegraphics[scale=0.05]{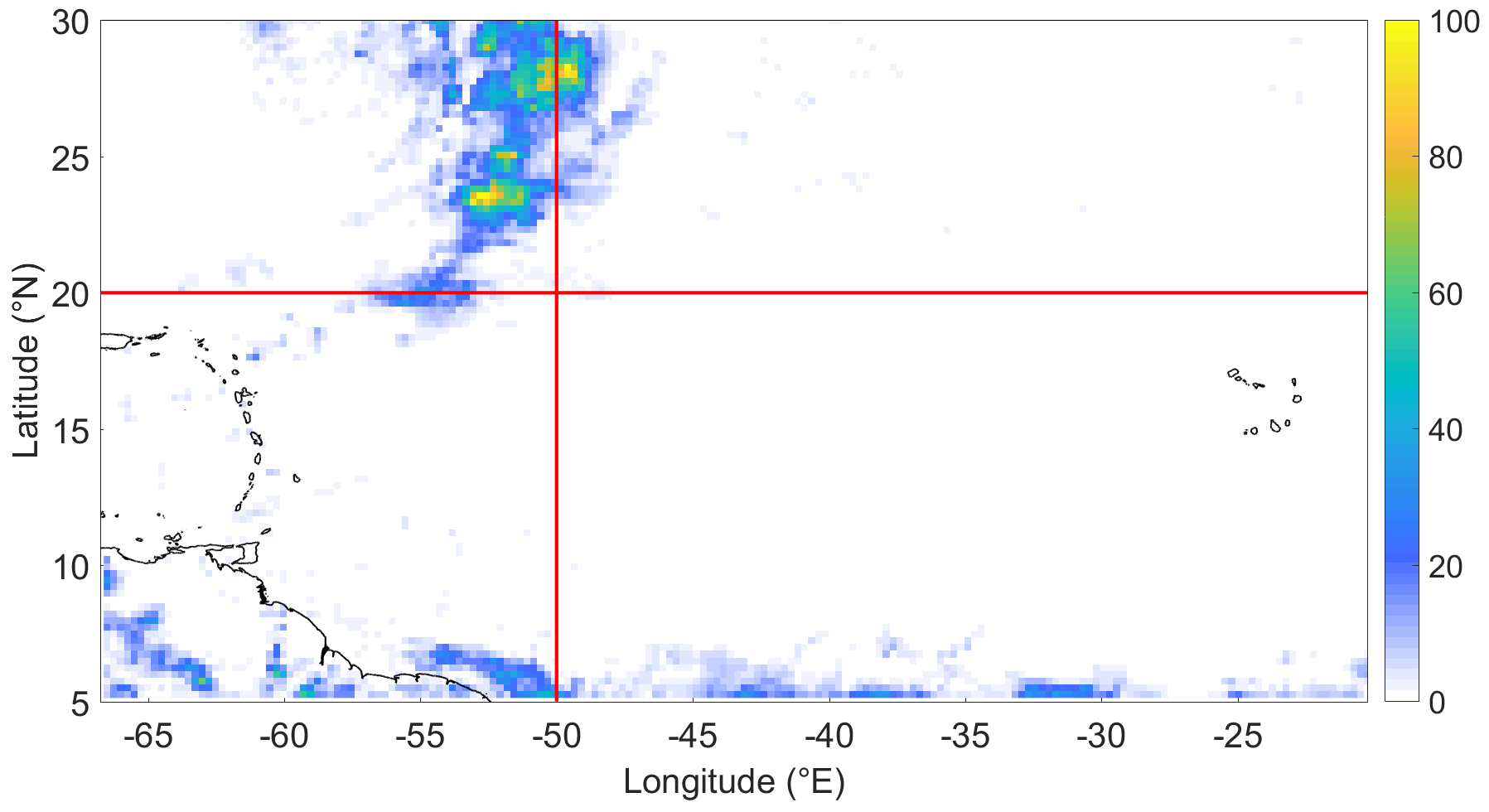}
\includegraphics[scale=0.05]{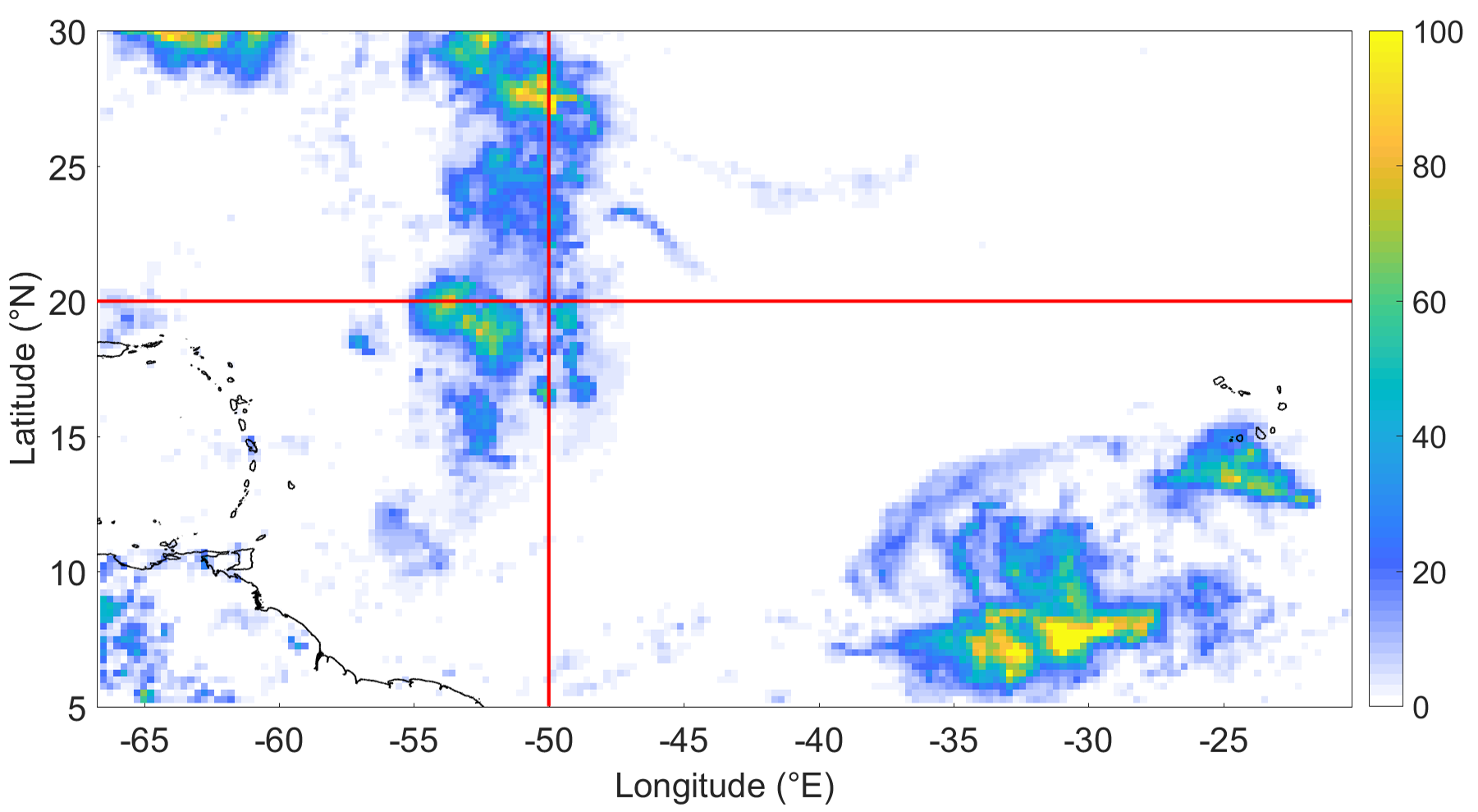}
\includegraphics[scale=0.05]{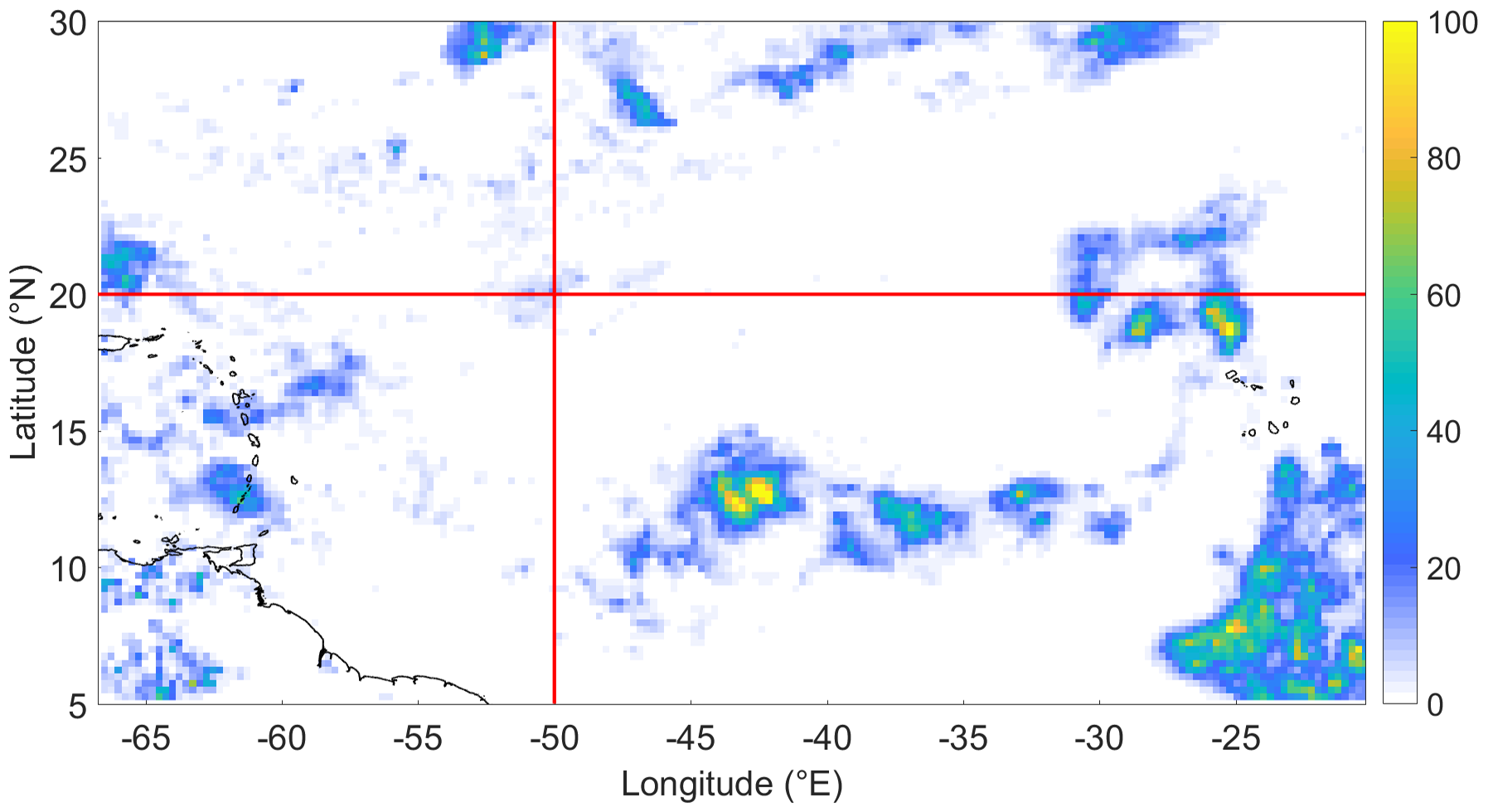}
\includegraphics[scale=0.05]{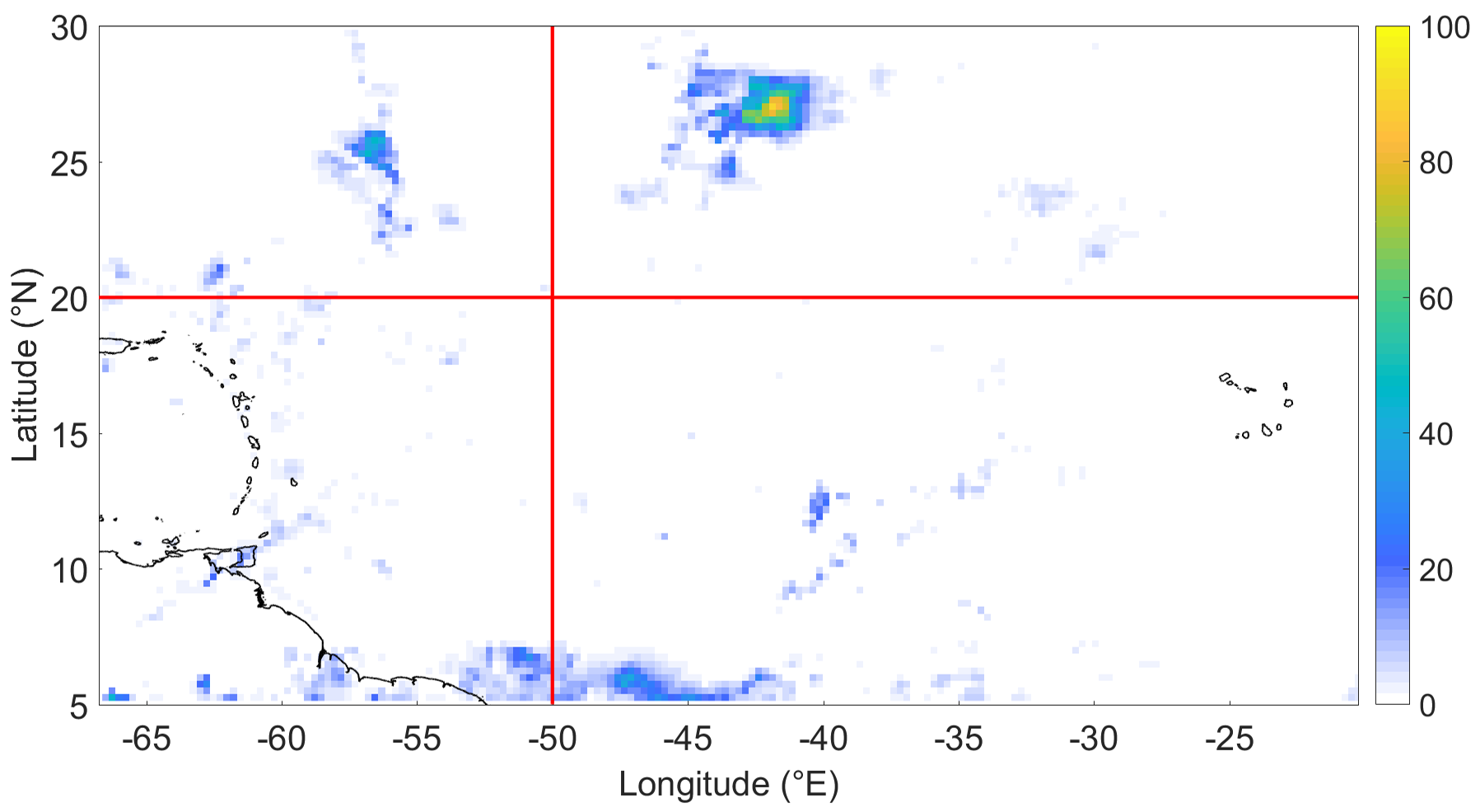}
\includegraphics[scale=0.05]{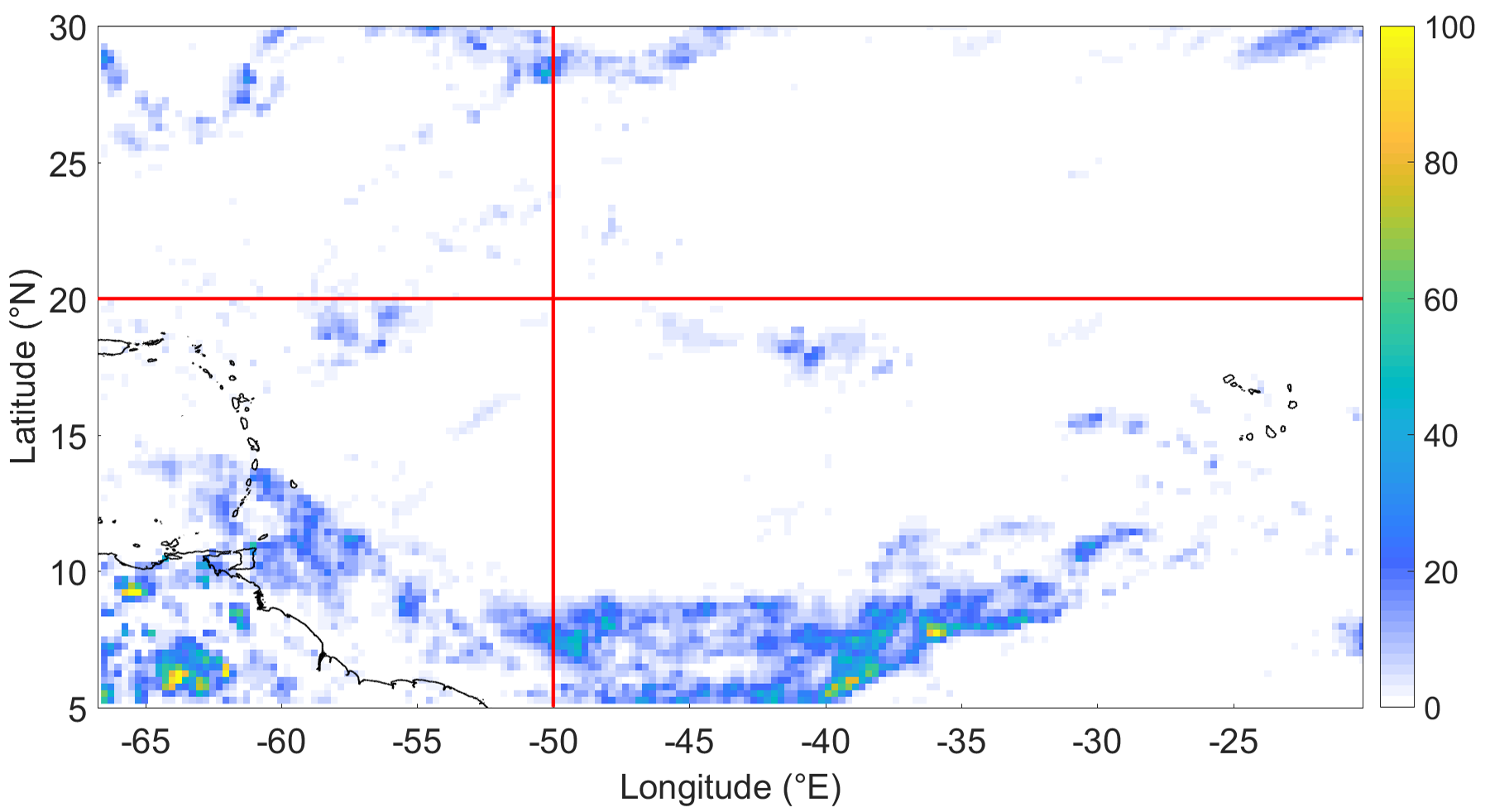}\\
(b)\includegraphics[scale=0.05]{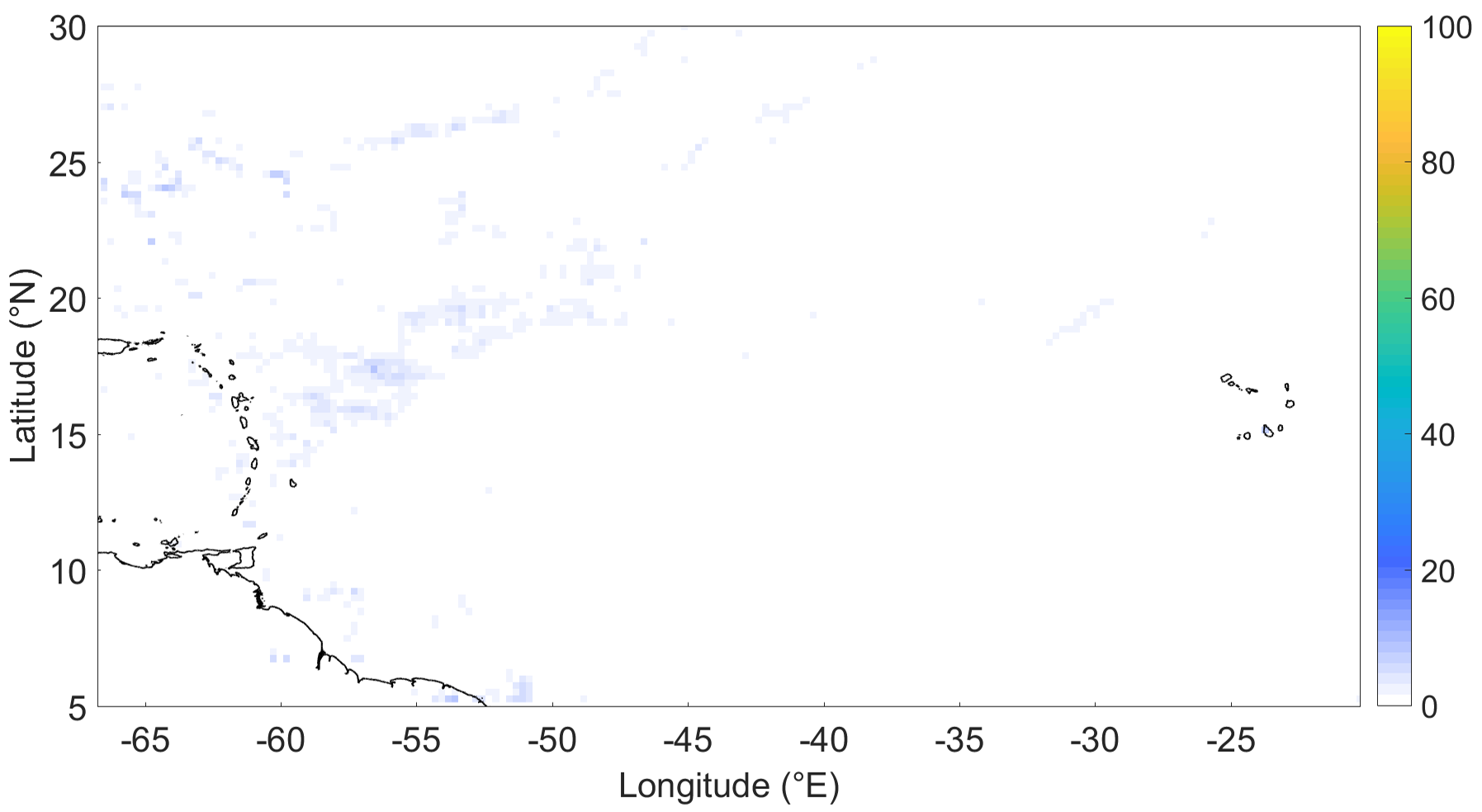}
\includegraphics[scale=0.05]{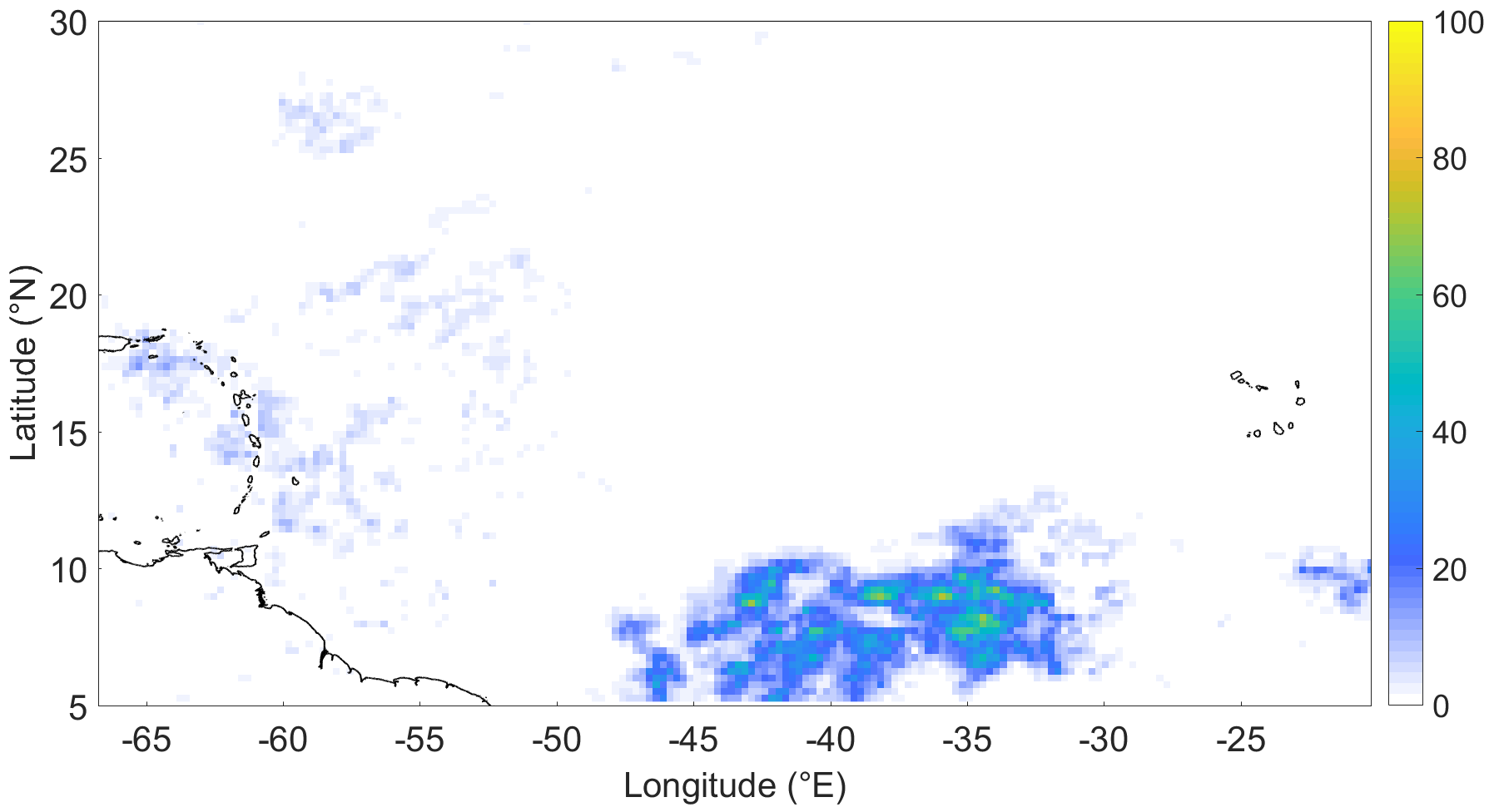}
\includegraphics[scale=0.05]{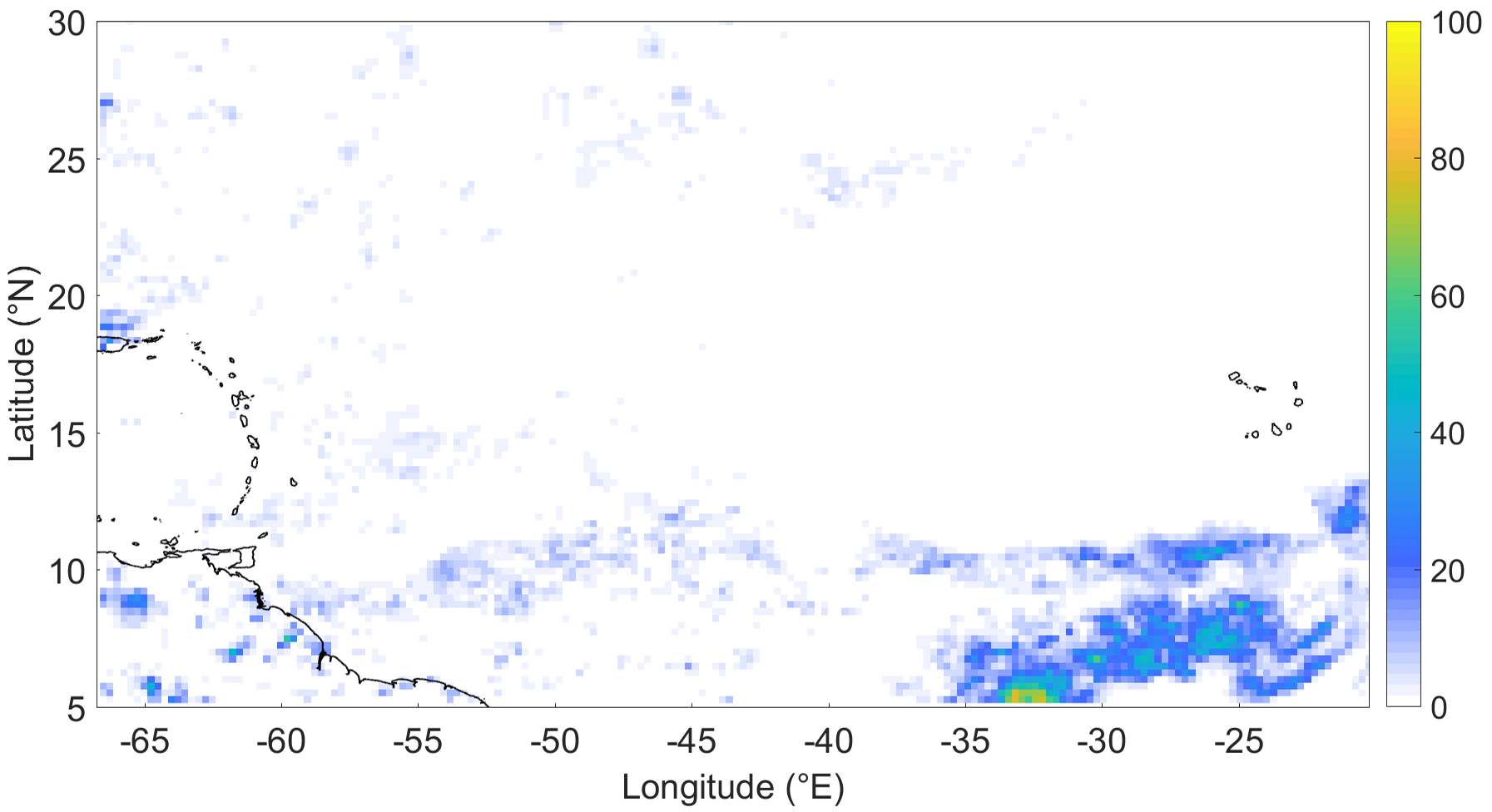}
\includegraphics[scale=0.05]{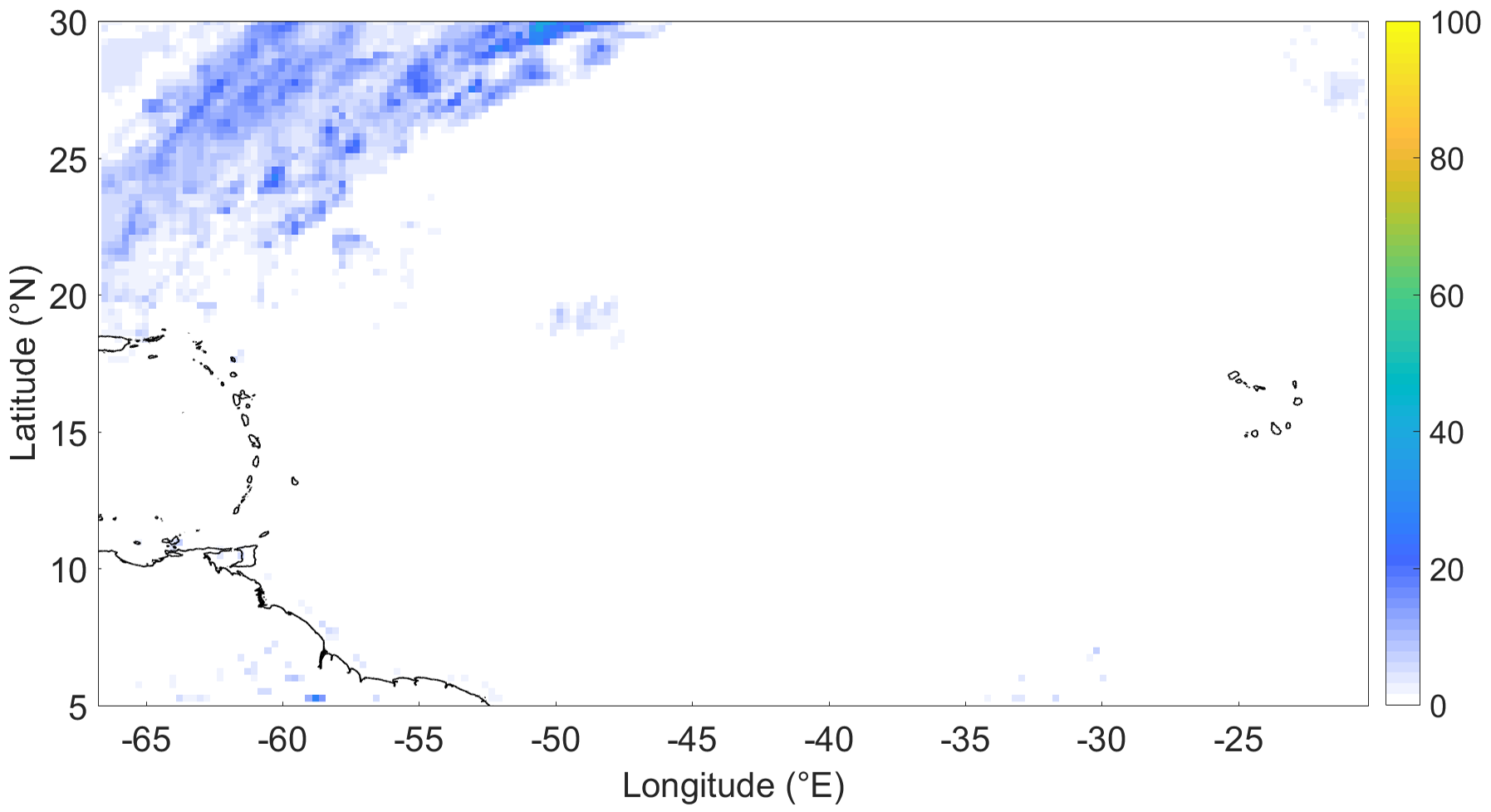}
\includegraphics[scale=0.05]{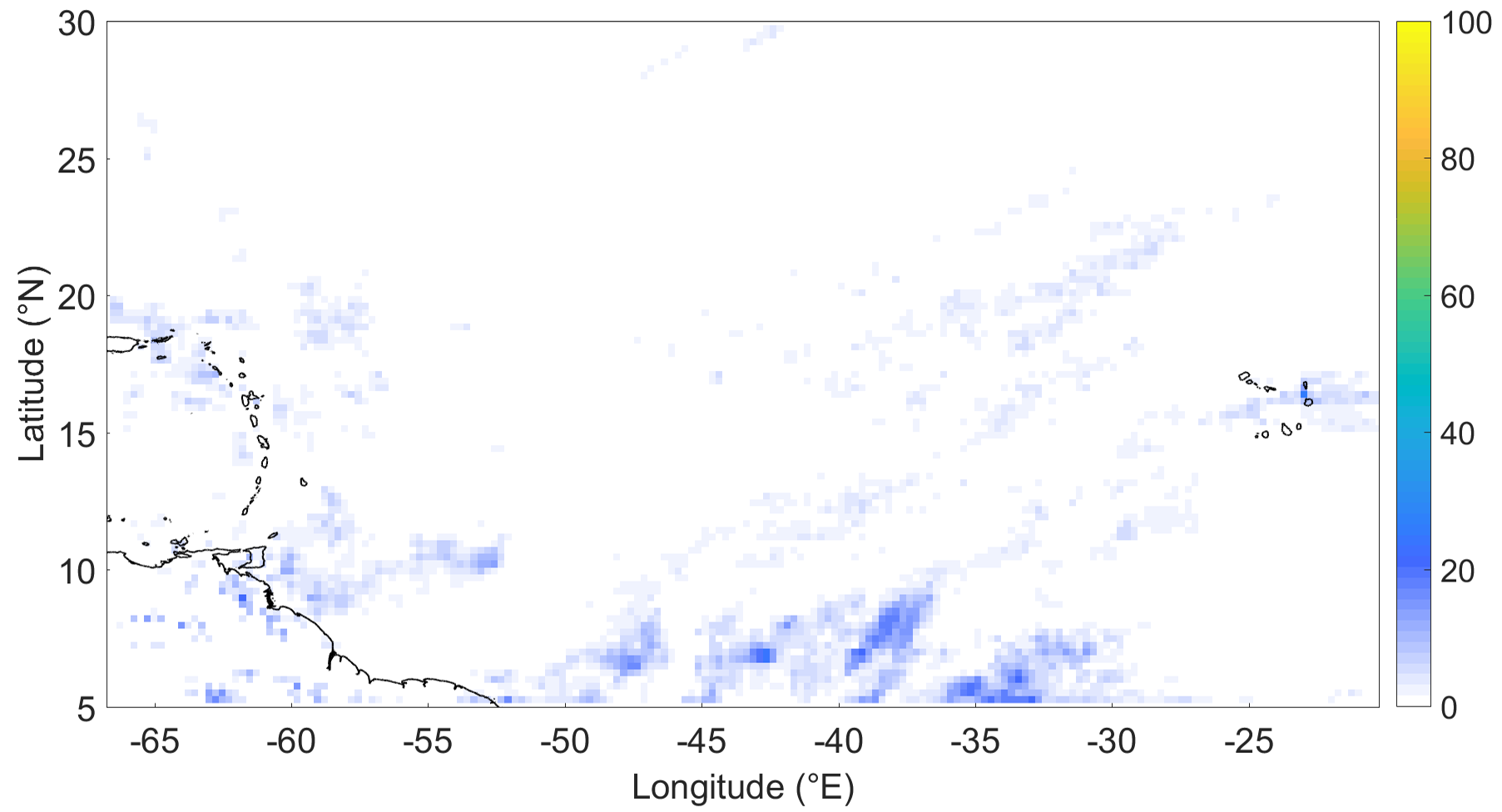}
\caption{Comparative graph of representative elements of the clusters of surface cumulative rainfall from the KMS-ED$_{RAINFALL}$ (a) and KMS-L2$_{RAINFALL}$ (b) method, with $k = 5$.}
\label{fig:sch_rain_clus}
\end{figure*}

\begin{figure*}
\centering
(a)\includegraphics[scale=0.065]{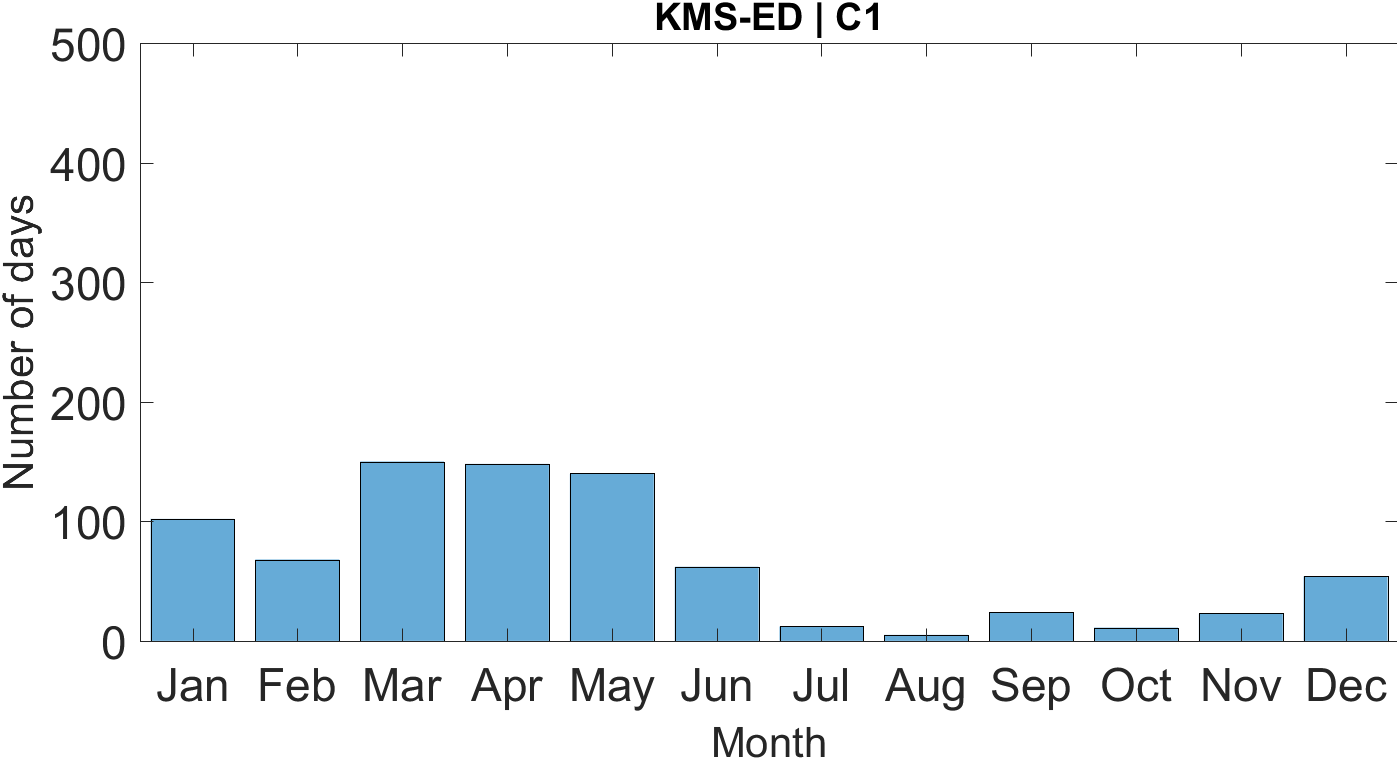}
\includegraphics[scale=0.065]{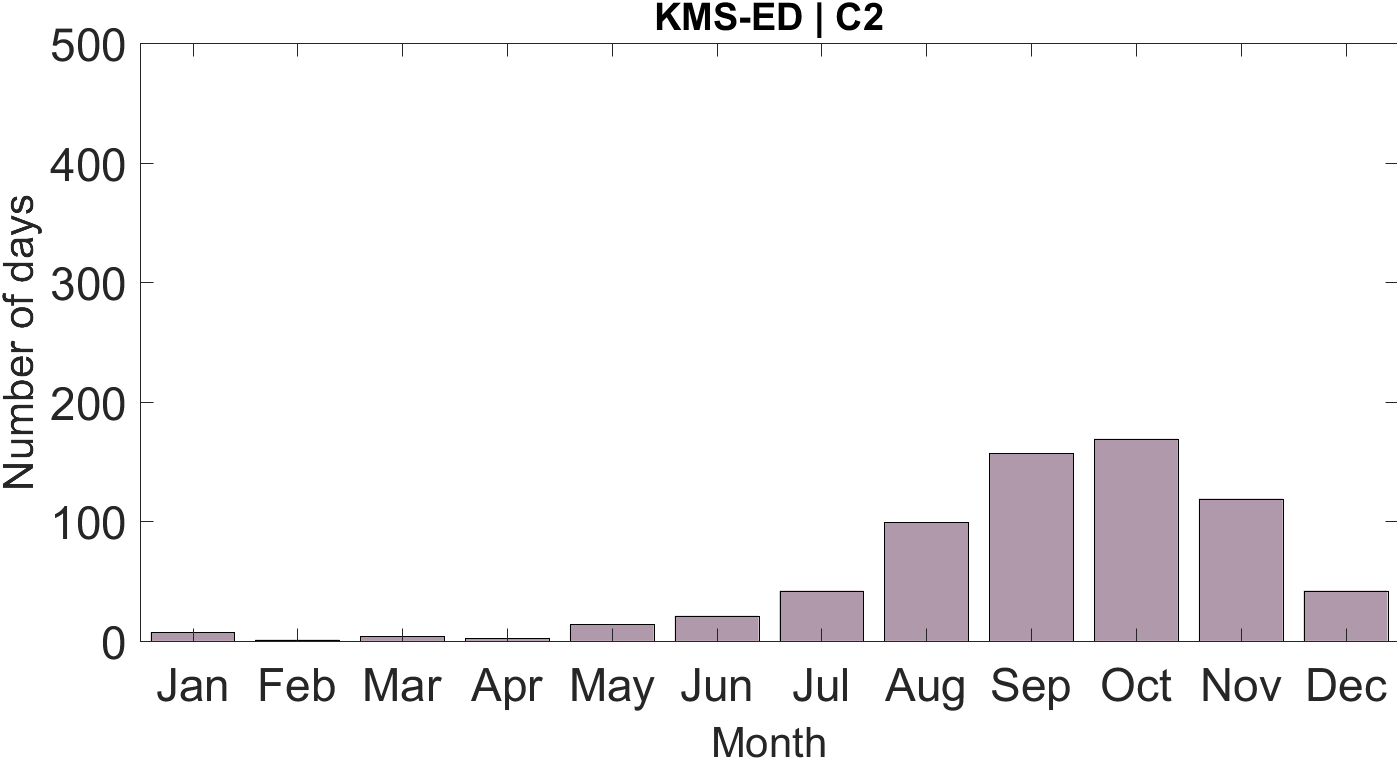}
\includegraphics[scale=0.065]{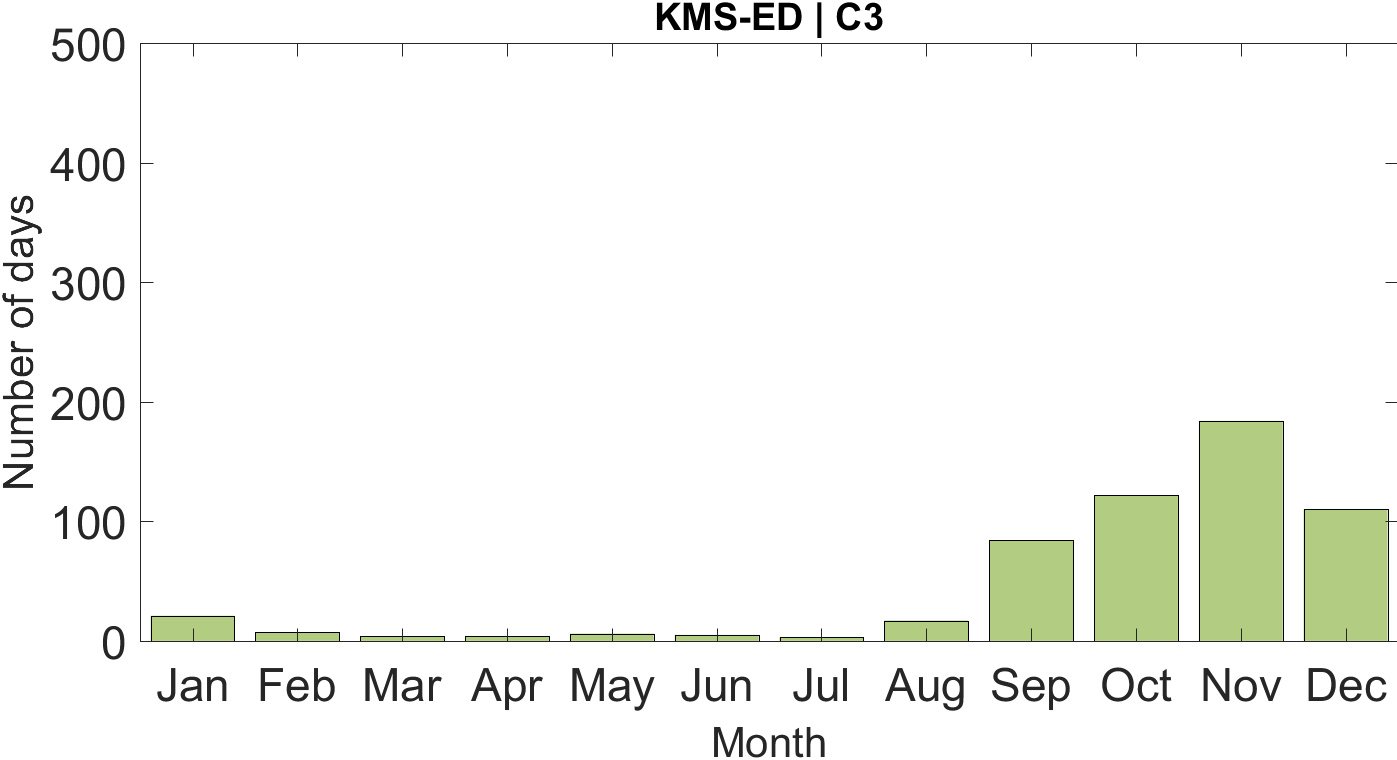}
\includegraphics[scale=0.065]{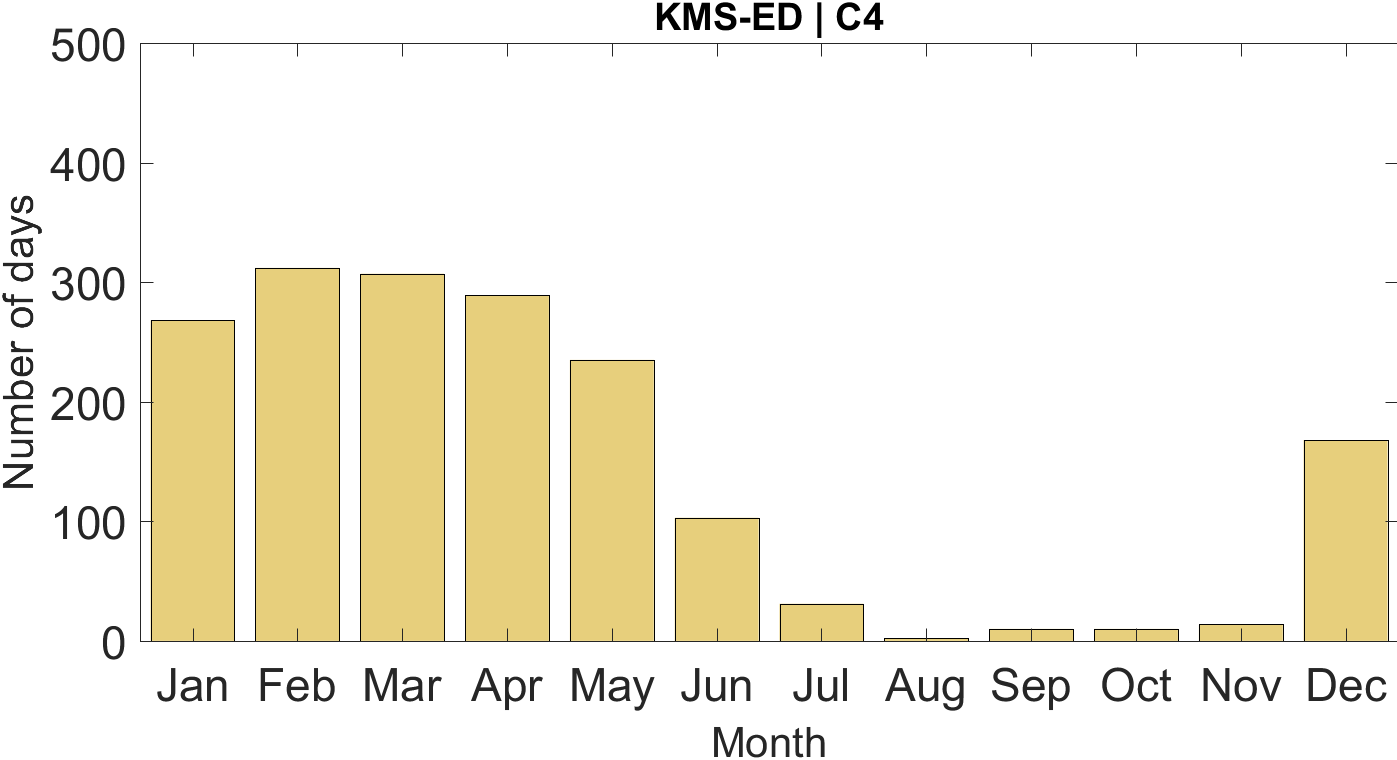}
\includegraphics[scale=0.065]{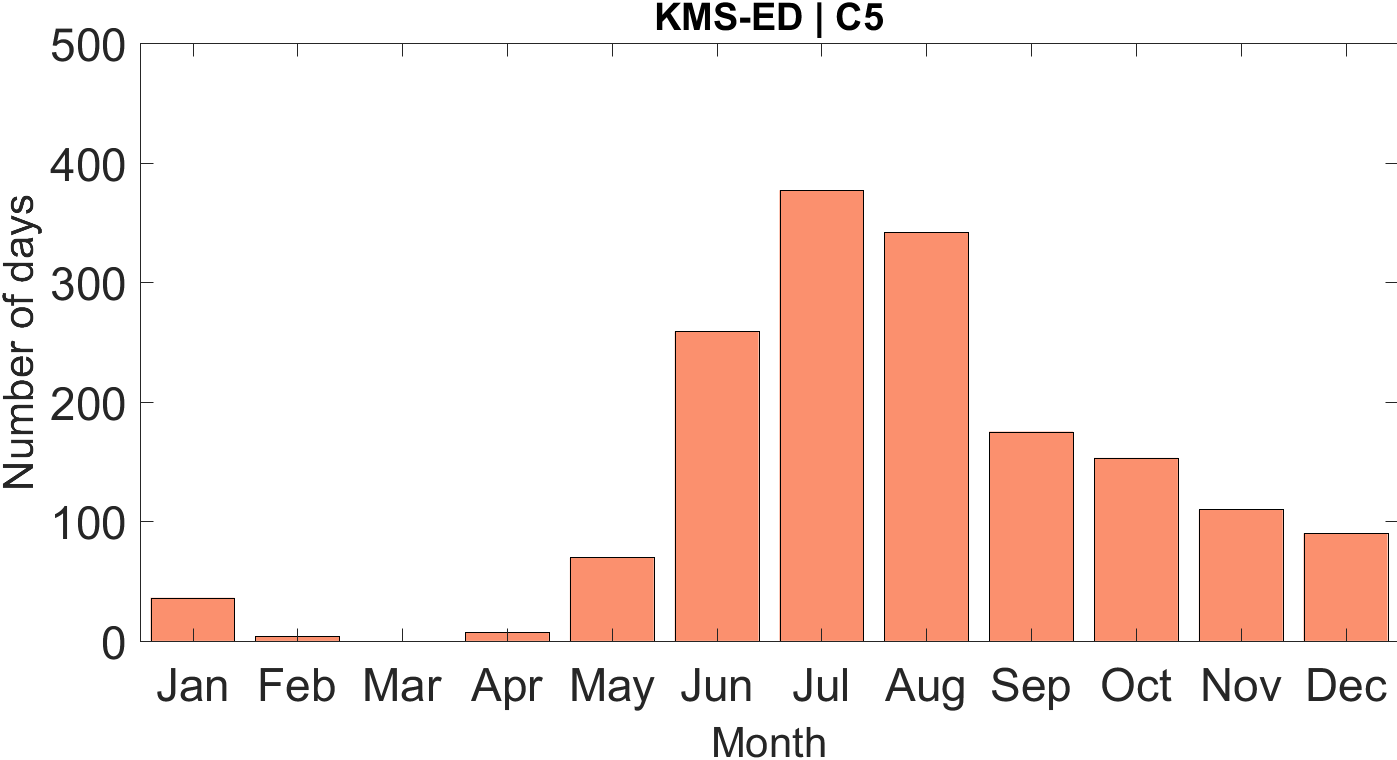}\\
(b)\includegraphics[scale=0.065]{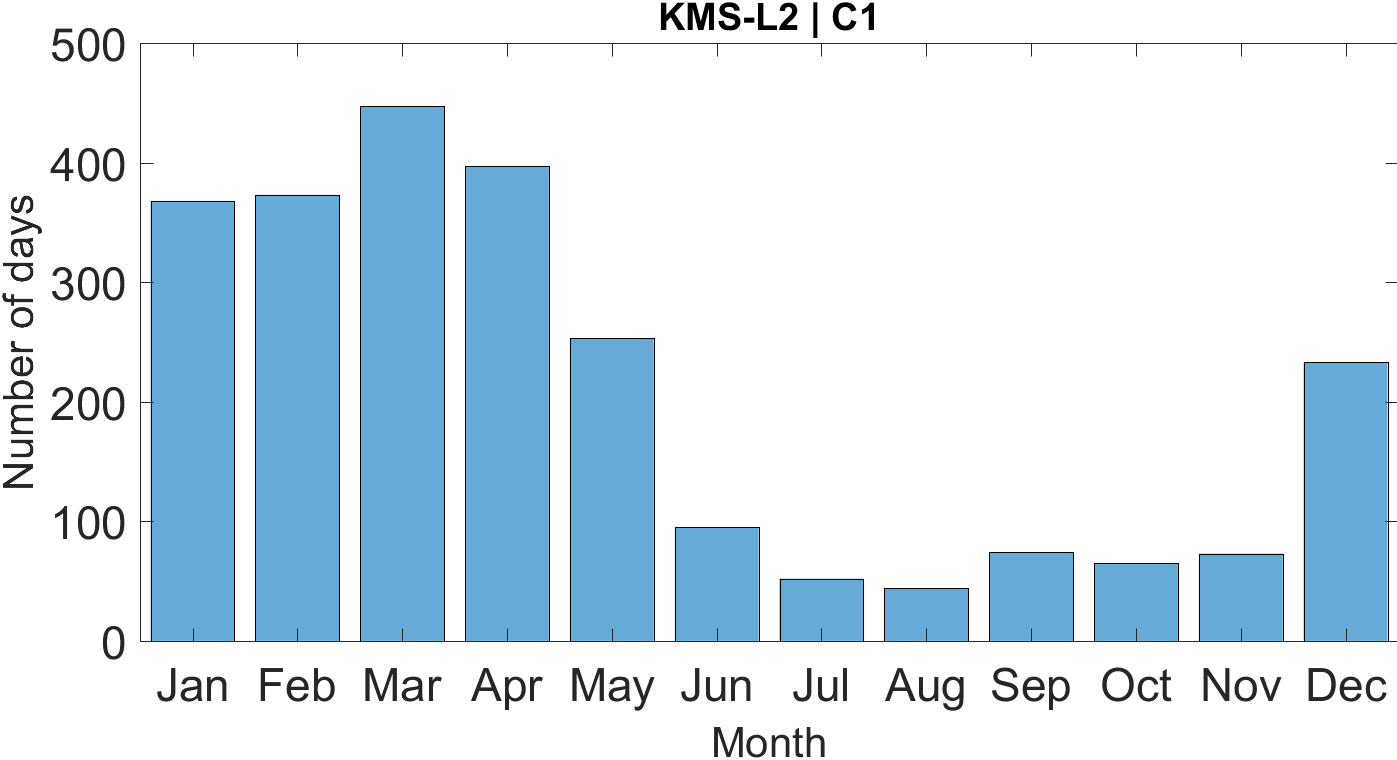}
\includegraphics[scale=0.065]{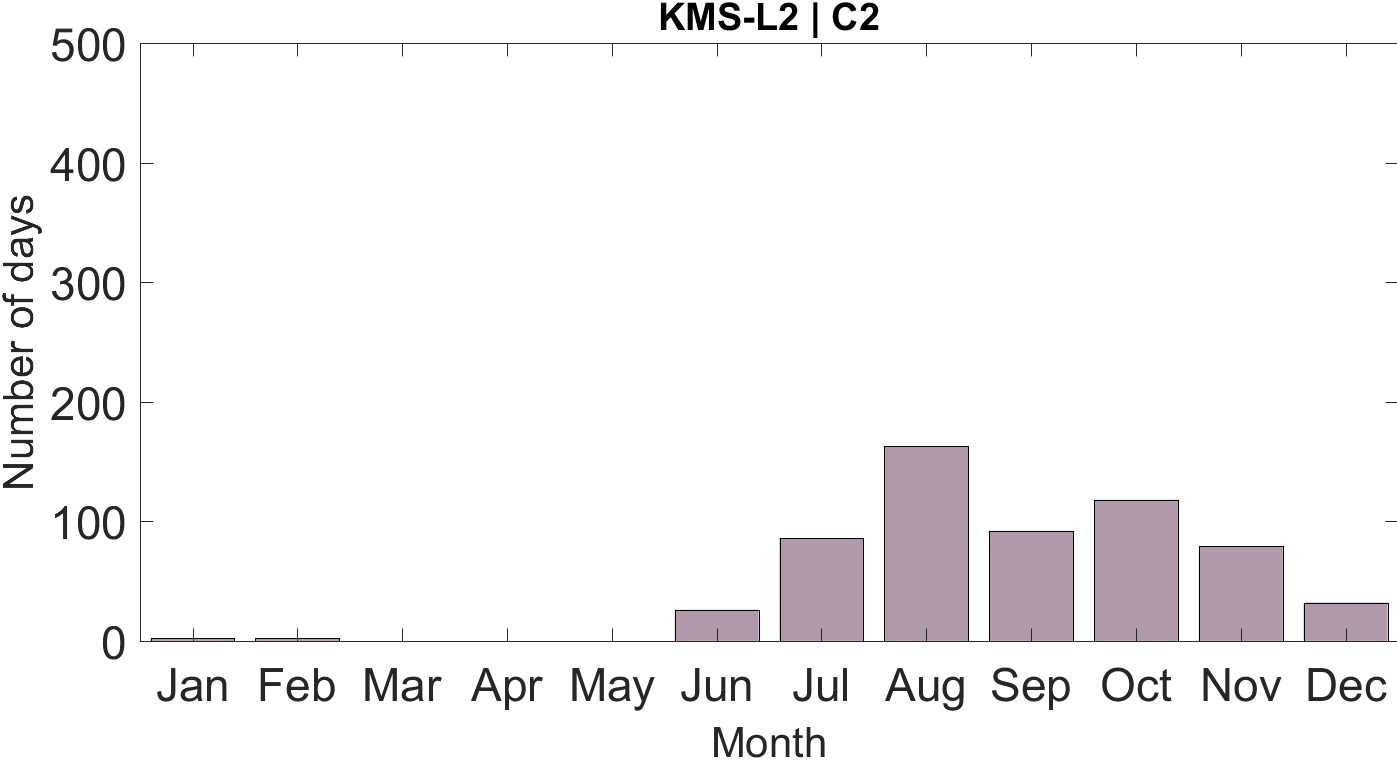}
\includegraphics[scale=0.065]{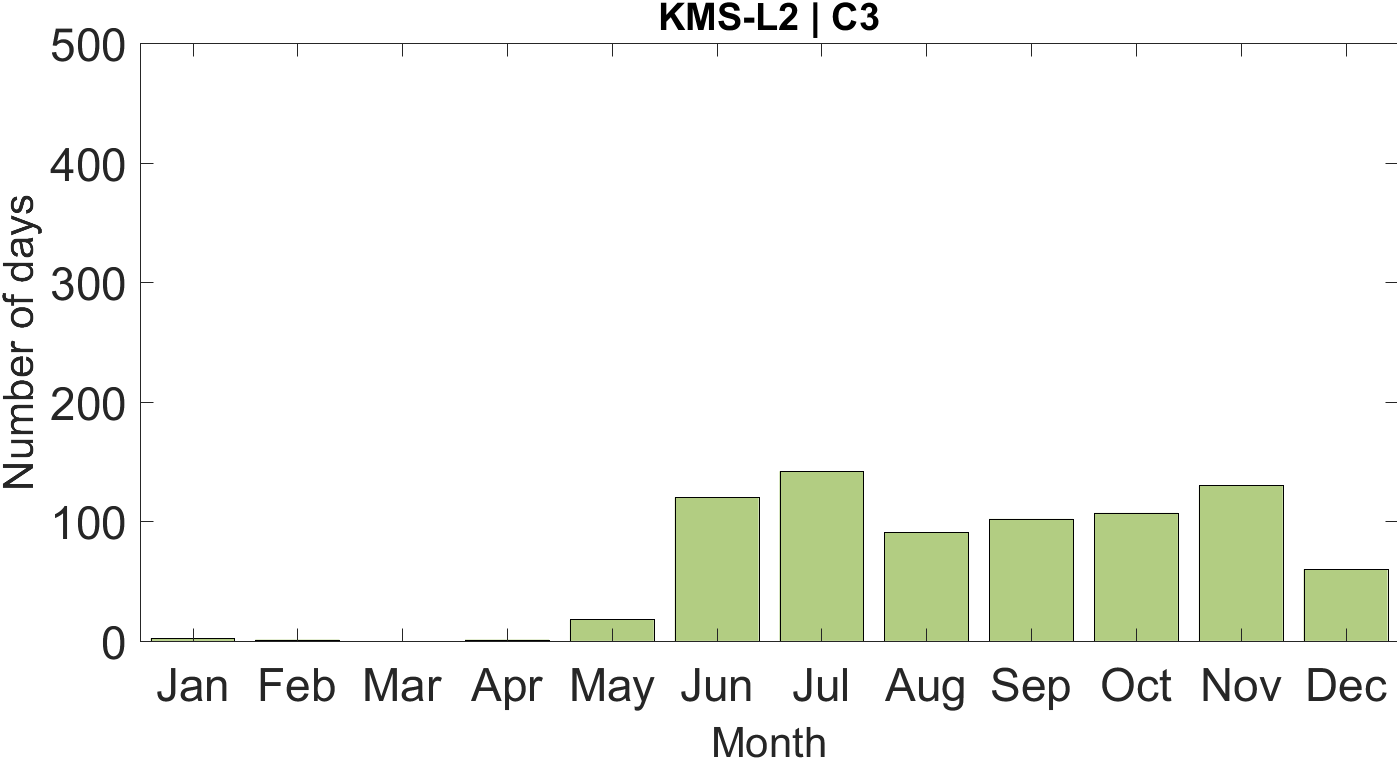}
\includegraphics[scale=0.065]{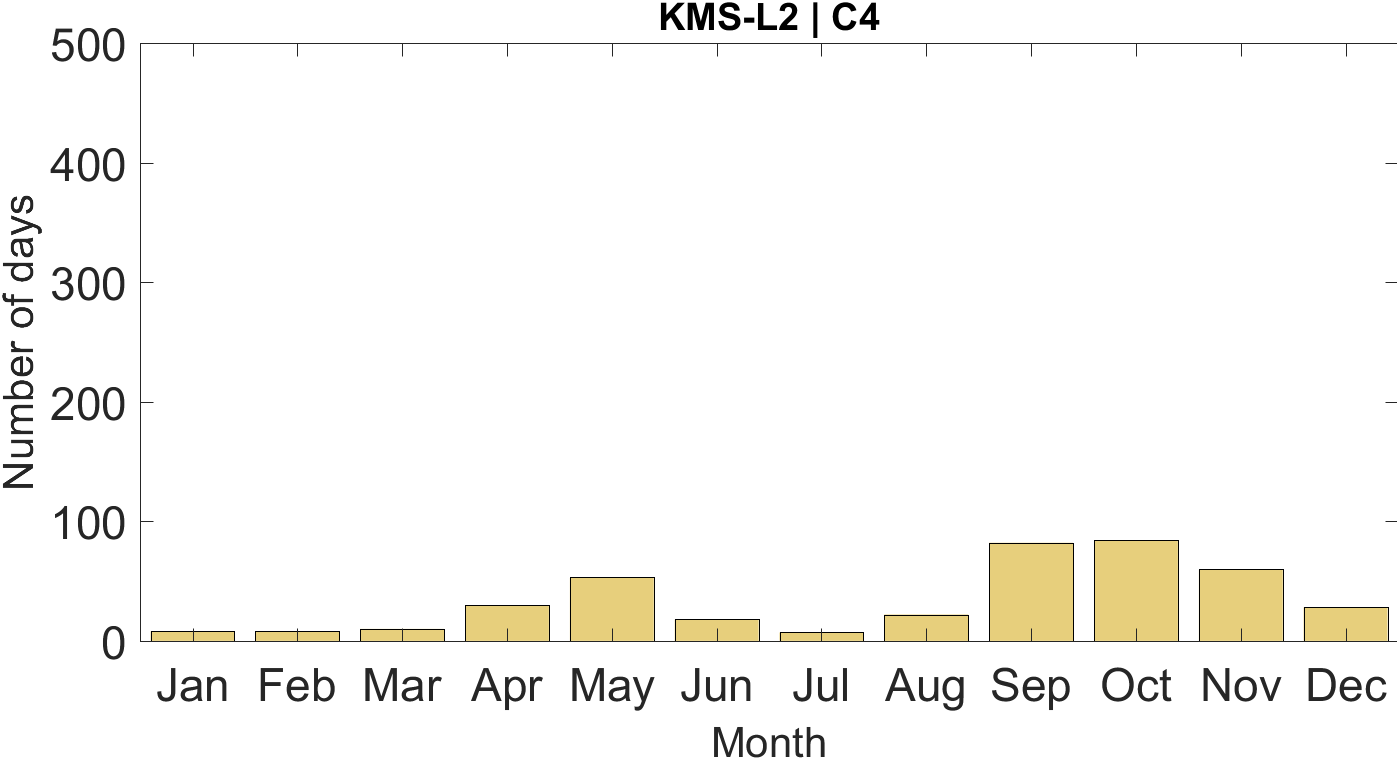}
\includegraphics[scale=0.065]{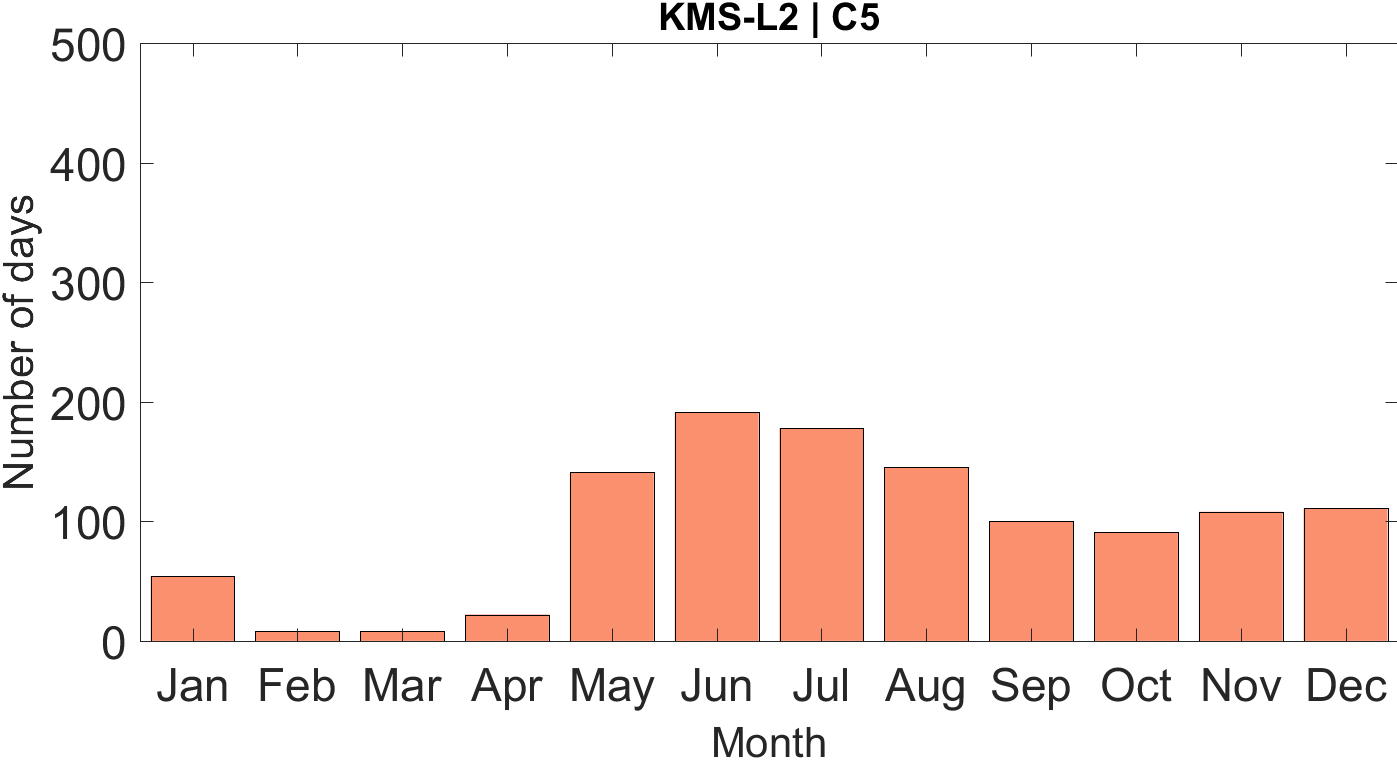}
\caption{Monthly distribution of clusters using the KMS-ED$_{RAINFALL}$ (a) and KMS-L2$_{RAINFALL}$ (b) method over the year for the period 2000 to 2014.}
\label{fig:sch_rain_dist}
\end{figure*}

\noindent The interest of our approach is reinforced by figures \ref{fig:sch_rain_clus} and \ref{fig:sch_rain_dist} which present the same studies on the second dataset (cumulative rainfalls). According to the experts, the results are equally significant since the representative elements of KMS-ED$_{RAINFALL}$ describe different configurations of daily cumulative rainfall (Fig \ref{fig:sch_rain_clus}a) whereas KMS-L2$_{RAINFALL}$ ones show dry situations with a active ITZC for certain clusters and cold surges arrival form north-west for another (Fig \ref{fig:sch_rain_clus}b). 

\noindent Experts highlight that monthly distribution of clusters C4 and C5 of KMS-DE$_{RAINFALL}$ respectively correspond to dry and rainy periods for the Lesser Antilles. C1, C2 and C3 are in the transition periods between the two main seasons (Fig \ref{fig:sch_rain_dist}a). For KMS-L2$_{RAINFALL}$, the monthly distribution of C1 corresponds to dry season where C2, C3 and C5 are  in rainy season. C4 seems to represent cold surge situation which appends during the year (Fig \ref{fig:sch_rain_dist}b).

\noindent This evaluation of the clusters by experts of the application field confirms the numerical results from the silhouette index, for every dataset tested: The Expert Deviation is by far a better candidate for clustering meteorological data than L2.

\section{Discussions}
\label{sec:discuss}
In this article, clustering was used to analyze the characteristics of the spatio-temporal fields of the cumulative rainfall obtained by satellite measurements on either side of the Lesser Antilles (the Atlantic Ocean and the Caribbean Sea). Our aim was to give more relevance and weight to physical information from clustering. We have shown that the L2 standard, frequently used in the classification of climate data, leads to groupings that can be made up of spatio-temporal fields having a very different interpretation or physical meaning.\\
\noindent The analysis we have carried out using these methods reveals the following difficulties: as rainfall totals are widely scattered, ``near'' fields in the meaning of L2 are rare; this norm tends to agglomerate fields with a common spatial structure, although they are otherwise different. The centroid used by KMS does not always represent a realistic physical situation.\\
\noindent It gives too much importance to data matching. This hinders the identification and interpretation of main trends in a region known for its high variability (Section \ref{subsec:prob_l2}). These difficulties have been confirmed by several publications such \cite{articlePB_L2,article77,article37}.
\noindent The innovation concerns the integration of a new metric, called expert deviation, to better quantify the similarity of the fields between them. This is a new step forward in this domain. Similar concerns have already emerged in the field of Image Retrieval, where surveys can be found in \cite{Rui1999,article72}. In addition, to increase the relevance of our results, we believed it was logical and judicious to develop a suitable subdivision of the fields (as suggested by \cite{mehtre1997,Manouvrier2005}).\\ 
\noindent So we spatially subdivided the daily fields into patches corresponding to known precipitation drivers. Thus, it allows us to include physical meaning in the proposed measure. 
Once these zones have been defined, the dissimilarity measure implemented will be based on a relaxation of the precision of the spatial location of the fields. 
\noindent To this end, it seemed logical to us to use measures related to information theory, such as the Kullback-Leibler divergence \cite{article64,article63}. Other measures of dissimilarity between histograms exist \cite{article77,article76}, but this latter was chosen because there was no a priori distribution of the data. From the patches of the daily rainfall fields, we compiled all histograms and compared them, using this divergence. The resulting method is still an unsupervised classification method. The innovation concerns the integration of an expert deviation to better quantify the similarity of the fields between them.\\
\noindent In spite of its originality, the Expert Deviation can be directly compared to conventional methods thanks to the silhouette index, usually used to quantify the clusters coherence (Fig \ref{fig:sch_silhouette}). These have been completed a visual interpretation of the clusters obtained for each method (Fig \ref{fig:sch_wind_clus} and \ref{fig:sch_rain_clus}).\\
\noindent From a numerical point of view, the use of ED produces systematically better results than L2. Thus, among the clusters obtained by KMS-ED, five of them were retained.
\noindent The design of this new metric is more responsive to the knowledges of the meteorological structures issued from observational data and scientific literature \cite{article5,article34,Stephenson14}.
\noindent It applies both to a complex and intermittent spatio-temporal field (rain) and to a classical one like winds. It is difficult to compare our results with those of other authors because few studies have combined this type of measurement with a regional subdivision of the field to be analyzed.\\
\noindent However, the proposed ED provides more reliable and interpretable results compared to traditional clustering methods.  This is the most effective way to produce results that are truly reliable and interpretable. It requires a prior spatial separation and quantification of the field to improve the physical relevance of clusters and thus reveal possible changes in transitions between seasons.

\section{Conclusion and perspectives}
\label{sec:clc_pers}
Studying the climate using clustering methods is a very interesting approach since there is a large amount of data available which is constantly increasing over the years. However, it should be noted that the most common methods of analysis are not necessarily appropriate to all types of data and that these types of methods should be adjusted. Especially if the data have a spatio-temporal nature, such as meteorological parameters. \\
\noindent In this article, we proposed the use of a silhouette index to evaluate numerically the quality of the clusters. We raised some theoretic issues that might appear when using L2 as a distance for clustering meteorological data. We also proposed a new dissimilarity measure (ED) to overcome these issues. As expected, the new metric gives much better results than the traditional L2. Moreover, these results have been analyzed by atmospheric physicists who confirmed the quality of the information that can be extracted from these clusters.\\
\noindent Unlike L2, the ED is able to produce different configurations which renders the usual atmospheric structures clearly identifiable. Atmospheric physicists can interpret impacts of each cluster on a specific zone according to atmospheric structure locations. KMS-L2$_{WIND}$ does not provide this, the situations represented are spatially quite smooth, the usual structures are not clearly visible.\\

\noindent The perspectives of this work are multiple: At first, one could design expert deviations for each meteorological parameter characterizing the climate. The use of a numerical evaluation now allows comparison of the quality of the discovered clusters for each possible parameter. The analysis of the dynamic of the clusters over time can also be an interesting way to continue this work.\\ 
\noindent It should also be possible to design expert deviation in order to cluster the days according to several parameters. A clustering with this combined ED could allow a more complete comparison of days since it would depend on multiple meteorological variables.\\
\noindent At last, it would be very interesting to cluster data according to a parameter (let us say the wind) and see if the days grouped into clusters are relevant when observed from a second parameter (rainfall for example), hence enabling to define which parameter influences the most to other observations.

\bibliographystyle{plain}
\bibliography{ReferencesArticles}

\begin{thebibliography}{10}

\bibitem{articlePB_L2}
Charu~C. Aggarwal, Alexander Hinneburg, and Daniel~A. Keim.
\newblock On the surprising behavior of distance metrics in high dimensional
  space.
\newblock In Jan Van~den Bussche and Victor Vianu, editors, {\em Database
  Theory --- ICDT 2001}, pages 420--434, Berlin, Heidelberg, 2001. Springer
  Berlin Heidelberg.

\bibitem{article63}
Alessia Amelio and Clara Pizzuti.
\newblock A patch-based measure for image dissimilarity.
\newblock {\em Neurocomputing}, 171:362--378, 2016.

\bibitem{article65}
Connelly Barnes, Dan~B. Goldman, Eli Shechtman, and Adam Finkelstein.
\newblock The patchmatch randomized matching algorithm for image manipulation.
\newblock {\em Commun. ACM}, 54(11):103--110, November 2011.

\bibitem{Boltz2007}
S.~{Boltz}, E.~{Debreuve}, and M.~{Barlaud}.
\newblock High-dimensional statistical distance for region-of-interest
  tracking: Application to combining a soft geometric constraint with
  radiometry.
\newblock In {\em 2007 IEEE Conference on Computer Vision and Pattern
  Recognition}, pages 1--8, June 2007.

\bibitem{article59}
M.~Burlando.
\newblock The synoptic-scale surface wind climate regimes of the mediterranean
  sea according to the cluster analysis of era-40 wind fields.
\newblock {\em Theoretical and Applied Climatology}, 96(1):69--83, April 2009.

\bibitem{article76}
Sung-Hyuk Cha and Sargur~N. Srihari.
\newblock On measuring the distance between histograms.
\newblock {\em Pattern Recognition}, 35(6):1355 -- 1370, 2002.

\bibitem{article34}
Xsitaaz~T. Chadee and Ricardo~M. Clarke.
\newblock Daily near-surface large-scale atmospheric circulation patterns over
  the wider caribbean.
\newblock {\em Climate Dynamics}, 44(11):2927--2946, Juin 2015.

\bibitem{article64}
Liviu~Petrisor Dinu, Radu-Tudor Ionescu, and Marius Popescu.
\newblock Local patch dissimilarity for images.
\newblock In Tingwen Huang, Zhigang Zeng, Chuandong Li, and Chi~Sing Leung,
  editors, {\em Neural Information Processing}, pages 117--126, Berlin,
  Heidelberg, 2012. Springer Berlin Heidelberg.

\bibitem{article69}
Alison~L. Gibbs and Francis~Edward Su.
\newblock On choosing and bounding probability metrics.
\newblock {\em International Statistical Review}, 70(3):419--435, 2002.

\bibitem{article62}
G.~Guo and C.~R. Dyer.
\newblock Patch-based image correlation with rapid filtering.
\newblock In {\em 2007 IEEE Conference on Computer Vision and Pattern
  Recognition}, pages 1--6, June 2007.

\bibitem{article68}
S.~Kullback and R.~Leibler.
\newblock On information and sufficiency.
\newblock {\em Annals of Mathematical Statistics}, 22:79--86, 1951.

\bibitem{article67}
R.~{Kwitt} and A.~{Uhl}.
\newblock Image similarity measurement by kullback-leibler divergences between
  complex wavelet subband statistics for texture retrieval.
\newblock In {\em 2008 15th IEEE International Conference on Image Processing},
  pages 933--936, Oct 2008.

\bibitem{Manouvrier2005}
Maude Manouvrier, Marta Rukoz, and Geneviève Jomier.
\newblock A generalized metric distance between hierarchically partitioned
  images.
\newblock {\em Proceedings 6th Intl. Workshop on Multimedia Data Mining -
  "Mining Integrated Media and Complex Data". MDM/KDD2005 -Chicago (USA). In
  conjunction with the Eleventh ACM SIGKDD Int. Conf. on Knowledge Discovery
  and Data Mi}, pages 33--41, 08 2005.

\bibitem{article36}
Carlos Martinez, Lisa Goddard, Yochanan Kushnir, and Mingfang Ting.
\newblock Seasonal climatology and dynamical mechanisms of rainfall in the
  caribbean.
\newblock {\em Climate Dynamics}, Jan 2019.

\bibitem{mehtre1997}
Babu~M. Mehtre, Mohan~S. Kankanhalli, and Wing~Foon Lee.
\newblock Shape measures for content based image retrieval: A comparison.
\newblock {\em Information Processing and Management}, 33(3):319 -- 337, 1997.

\bibitem{article4}
C.~Monteleoni, G.~A. Schmidt, F.~Alexander, A.~Niculescu-Mizil,
  K.~Steinhaeuser, M.~Tippett, A.~Banerjee, M.~B. Blumenthal, A.~R. Ganguly,
  J.~E. Smerdon, and M.~Tedesco.
\newblock {\em Climate informatics}, pages 81--126.
\newblock Data Mining and Knowledge Discovery Series. Chapman and Hall/CRC,
  2013.

\bibitem{article5}
Vincent Moron, Isabelle Gouirand, and Michael Taylor.
\newblock Weather types across the caribbean basin and their relationship with
  rainfall and sea surface temperature.
\newblock {\em Climate Dynamics}, 47(1):601--621, Jul 2016.

\bibitem{article77}
I.~Olkin and F.~Pukelsheim.
\newblock The distance between two random vectors with given dispersion
  matrices.
\newblock {\em Linear Algebra and its Applications}, 48:257 -- 263, 1982.

\bibitem{article48}
Shraddha Pandit and Suchita Gupta.
\newblock A comparative study on distance measuring approches for clustering.
\newblock {\em International Journal of Research in Computer Science},
  2:29--31, 2011.

\bibitem{article29}
Peter~J. Rousseeuw.
\newblock Silhouettes: a graphical aid to the interpretation and validation of
  cluster analysis.
\newblock {\em Computational and Applied Mathematics}, 20:53–65, 1987.

\bibitem{Rui1999}
Yong Rui, Thomas~S. Huang, and Shih-Fu Chang.
\newblock Image retrieval: Current techniques, promising directions, and open
  issues.
\newblock {\em J. Visual Communication and Image Representation}, 10:39--62,
  1999.

\bibitem{article37}
Pandit Shraddha and Gupta Suchita.
\newblock Computer science and technology. computing. data processing.
\newblock {\em International journal of research in computer science}, pages
  29--31, 2011.

\bibitem{article72}
A.~W.~M. {Smeulders}, M.~{Worring}, S.~{Santini}, A.~{Gupta}, and R.~{Jain}.
\newblock Content-based image retrieval at the end of the early years.
\newblock {\em IEEE Transactions on Pattern Analysis and Machine Intelligence},
  22(12):1349--1380, December 2000.

\bibitem{Stephenson14}
Tannecia~S. Stephenson, Lucie~A. Vincent, Theodore Allen, Cedric~J.
  Van~Meerbeeck, Natalie McLean, Thomas~C. Peterson, Michael~A. Taylor,
  Arlene~P. Aaron-Morrison, Thomas Auguste, Didier Bernard, Joffrey R.~I.
  Boekhoudt, Rosalind~C. Blenman, George~C. Braithwaite, Glenroy Brown, Mary
  Butler, Catherine J.~M. Cumberbatch, Sheryl Etienne-Leblanc, Dale~E. Lake,
  Delver~E. Martin, Joan~L. McDonald, Maria Ozoria~Zaruela, Avalon~O. Porter,
  Mayra Santana~Ramirez, Gerard~A. Tamar, Bridget~A. Roberts, Sukarni
  Sallons~Mitro, Adrian Shaw, Jacqueline~M. Spence, Amos Winter, and Adrian~R.
  Trotman.
\newblock Changes in extreme temperature and precipitation in the caribbean
  region, 1961–2010.
\newblock {\em International Journal of Climatology}, 34(9):2957--2971, 2014.

\bibitem{article78}
Stephen Walker, Paul Damien, and Peter Lenk.
\newblock On priors with a kullback-leibler property.
\newblock {\em Journal of the American Statistical Association},
  99(466):404--408, 2004.

\bibitem{article70}
Li~Xin.
\newblock {\em Perceptual Digital Imaging: Methods and Applications}, volume~1,
  chapter Patch-Based Image Processing: From Dictionary Learning to Structural
  Clustering.
\newblock CRC Press, March 2017.

\bibitem{article13}
Qin Zhang, Huug Van Den~Dool, Suru Saha, Malaquias Peña, Emily Becker, Peitao
  Peng, and Jin Huang.
\newblock Preliminary evaluation of multi-model ensemble system for monthly and
  seasonal prediction.
\newblock In NOAA’s National~Weather Service, editor, {\em Science and
  Technology Infusion Climate Bulletin}, pages 124--131, 2011.

\end{thebibliography}

\end{document}